%% file: main.tex
\documentclass[10pt,twocolumn,letterpaper,dvipsnames]{article}

\usepackage{3dv}
\usepackage{times}
\usepackage{epsfig}
\usepackage{graphicx}
\usepackage{amsmath}
\usepackage{amssymb}

\usepackage{booktabs}
\usepackage{multirow, makecell}
\usepackage{subcaption}
\usepackage{float}
\usepackage{rotating}
\usepackage{stix}
\usepackage{textgreek}
\usepackage{mathtools}
\usepackage{multibib}

\DeclareMathOperator*{\argmax}{\arg\!\max}
\usepackage[pagebackref=true,breaklinks=true,letterpaper=true,colorlinks=true,citecolor=blue,bookmarks=false]{hyperref}

\newcommand{\cR}{\mathcal{R}}

\newcommand\customparagraph[1]{\vspace{0.4em}\noindent\textbf{#1}}

\threedvfinalcopy %

\def\threedvPaperID{332} %

\ifthreedvfinal\fi

\begin{document}

\title{Digging Into Self-Supervised Learning of Feature Descriptors}

\author{
    Iaroslav Melekhov\thanks{indicates equal contribution. Correspondence to: \small\texttt{iaroslav.melekhov@aalto.fi}}
    \and Zakaria Laskar\footnotemark[1]
    \and Xiaotian Li\\
    \small{Aalto University}\\
    \and Shuzhe Wang\\
    \and Juho Kannala
    }

\maketitle
\thispagestyle{empty}

\begin{abstract}
\input{sections/abstract}

\end{abstract}

\input{sections/introduction}
\input{sections/related_work}

\input{sections/method}

\input{sections/experiments}
\input{sections/ablation}
\input{sections/conclusion}

\clearpage
\appendix
\input{supp/supp_main}

{\small
\bibliographystyle{ieee_fullname}
\bibliography{shortstrings,references}
}

\end{document}

%% file: sections/abstract.tex
Fully-supervised CNN-based approaches for learning local image descriptors have shown remarkable results in a wide range of geometric tasks. However, most of them require per-pixel ground-truth keypoint correspondence data which is difficult to acquire at scale. To address this challenge, recent weakly- and self-supervised methods can learn feature descriptors from relative camera poses or using only synthetic rigid transformations such as homographies. In this work, we focus on understanding the limitations of existing self-supervised approaches and propose a set of improvements that combined lead to powerful feature descriptors. We show that increasing the search space from in-pair to in-batch for hard negative mining brings consistent improvement. To enhance the discriminativeness of feature descriptors, we propose a coarse-to-fine method for mining local hard negatives from a wider search space by using global visual image descriptors. We demonstrate that a combination of synthetic homography transformation, color augmentation, and photorealistic image stylization produces useful representations that are viewpoint and illumination invariant. The feature descriptors learned by the proposed approach perform competitively and surpass their fully- and weakly-supervised counterparts on various geometric benchmarks such as image-based localization, sparse feature matching, and image retrieval.

%% file: sections/introduction.tex
\input{assets/images/fig_pipeline}

\section{Introduction}~\label{sec:introduction}
Many geometric computer vision tasks require robust estimation of local descriptors such as image alignment~\cite{Balntas2017HPatches}, structure-from-motion~\cite{heinly2015_reconstructing_the_world,schoenberger2016sfm}, image retrieval~\cite{Gordo2017Retrieval,Radenovic2018TPAMI}. With the advent of deep learning models, robust local descriptors can be learnt with highly over-parameterized networks such as Convolutional Neural Networks (CNN). Consequently, much of the latest works~\cite{Balntas2016Localdesc,Detone18superpoint,Dusmanu2019D2Net,Mishchuk2017HardNet,Revaud2019R2D2} have since focused on designing appropriate datasets, labels and loss functions to learn useful local representations.

It is well-known that CNNs are data demanding models that require large amount of data with labelled supervision. While in many cases large amount of data can be obtained, structuring it in meaningful partitions or labels is a costly process. To counter this problem, existing approaches~\cite{Gordo2017Retrieval,Radenovic2016GEM} use SfM pipelines to generate labels. Despite the effectiveness, the process incurs significant computational burden. Furthermore, integrating new stream of data to the existing dataset requires additional iterations of the costly spatial descriptor matching process. The paradigm of self-supervised learning attempts to provide a solution by enabling label-free training of CNNs. Most self-supervised methods focus on auxiliary pretext tasks~\cite{doersch_2015,noroozi_2016,zhang_2016} such as image reconstruction -- generative modelling, or discriminative optimization of the latent space based on contrastive learning~\cite{bachman_2019,dosovitskiy_2014}. The key requirement is to learn transferable representations that can adapt to downstream tasks with limited supervision. Recent progress~\cite{chen2020SimCLR,debiased,he2020moco,kalantidis,robinson_2021} in the image classification domain shows the crucial role of positives (similar) and negatives (dissimilar) in the contrastive loss function to learn robust representations. In this paper, we delve deeper into these factors for the task of local image descriptors learning.

Existing self-supervised methods for learning local descriptors~\cite{Christiansen2019Unsuperpoint,Detone18superpoint} have achieved some success by mining positives and negatives from homography related image pairs (in-pair sampling). On the other hand, \textit{supervised} local descriptor learning methods such as HardNet~\cite{Mishchuk2017HardNet} show consistent improvement with in-batch negative sampling. In addition, image retrieval methods~\cite{Gordo2017Retrieval,Radenovic2016GEM} have demonstrated remarkable results by mining global negative descriptors from a large database of thousands of images. In this paper, we propose a scalable method for sampling hard-negative local features from a wide search space that operates in a coarse-to-fine fashion by leveraging %
image descriptors.

Learning keypoint descriptors that are invariant to illumination changes is a challenging task since it is not easy to acquire suitable data. Synthetic color augmentation (CA) which is widely used to improve model robustness and performance in many computer vision applications~\cite{Fang2019InstaBoost,krizhevsky2012Imagenet,mohan2020efficientps,wu2019detectron2} can enhance illumination invariance of local descriptors only to a certain extent. Following~\cite{Revaud2019R2D2}, we propose to utilize photorealistic image stylization~\cite{Li2018Stylization} at training time. In contrast to~\cite{Revaud2019R2D2}, the more diverse style images are considered which help to increase appearance variations of our training data. We combine these findings to train existing state-of-the-art local descriptor models without any labels. During training and evaluation, only a subset of local descriptors from dense descriptors produced by the network is used. The selection is based on local keypoint detectors such as SuperPoint~\cite{Detone18superpoint} providing efficient training iterations -- by limiting hard-negative mining only to a subset of local descriptors. The proposed method is compared with strong supervised and weakly-supervised baselines~\cite{Revaud2019R2D2,Wang2020CAPS} on a wide range of benchmarks such as visual localization, sparse feature matching, and image retrieval. The key findings are analyzed and complemented by in-depth ablation study.

In summary, this work makes the following contributions: (1) we  demonstrate that improving the quality of hard negative samples during training leads to more powerful and discriminative local keypoint descriptors; (2) by leveraging visual image descriptors, we propose a coarse-to-fine method for mining hard negative local descriptors at scale; (3) we investigate several CA techniques to improve the robustness of learned feature descriptors to illumination changes and show that photorealistic image stylizations yield best performance. We demonstrate that, together, these contributions largely close the gap between self-supervised and fully-supervised approaches for feature descriptor learning. %

%% file: assets/images/fig_pipeline.tex
\begin{figure*}[t!]
  \centering
    \includegraphics[width=\linewidth]{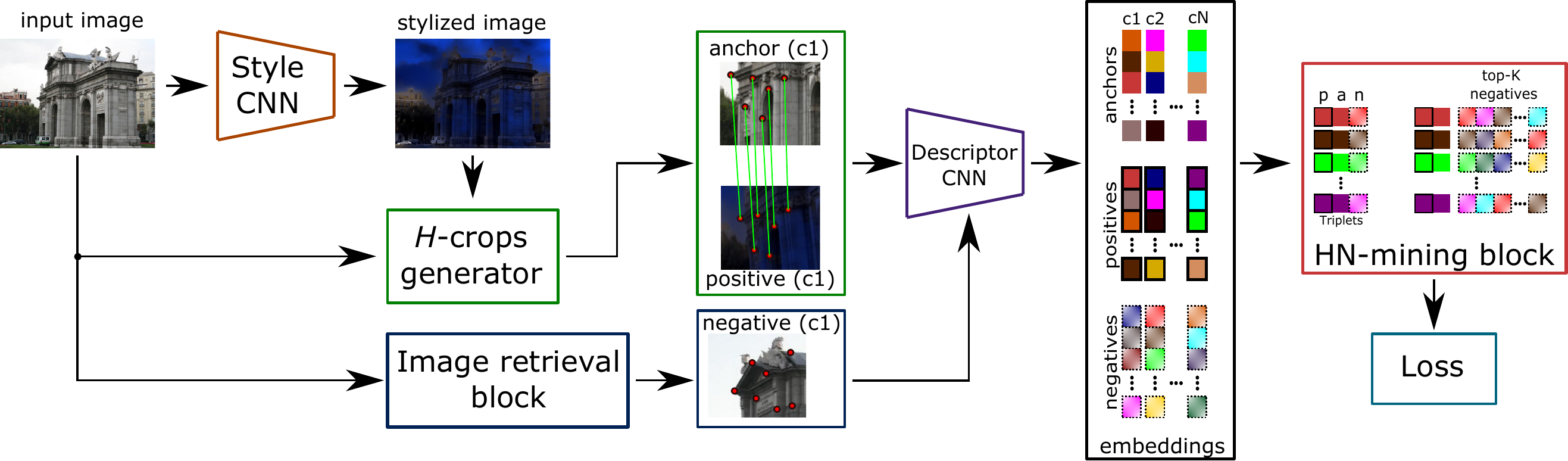}
     \vspace{-15pt}
 \caption{\textbf{Overview of the proposed pipeline for learning local image descriptors.} Given an image and a set of interest points (top-left), we create a stylized copy of the image with a forward-pass through Style CNN. A pair of homography related crops ($c_i$) is randomly sampled from the original and stylized image in H-crops generator. We pass the image crops through Descriptor CNN to obtain representation of corresponding interest points in the feature space (embeddings). For each $a_i$ and $p_i$ descriptor from positive pair (anchors-positives), we create a list of hard-negative descriptors (shown with color gradient) using the proposed coarse-to-fine sampling procedure by leveraging global visual descriptors (Image retrieval block). Finally, the ranking loss %
 is utilized to optimize Descriptor CNN. See Sec.~\ref{sec:method} for a detailed description.}  
 \label{fig:pipeline_overview}
 \vspace{-5pt}
\end{figure*}

%% file: sections/related_work.tex
\section{Related work}~\label{sef:related_work}
\hspace{-3mm}
\customparagraph{Local image descriptors.} Due to their generalization ability, traditional local features~\cite{Bay2006SURF,Calonder2010BRIEF,Lowe2004SIFT,Rublee2011ORB} are still popular and widely used by many existing systems where establishing pixel correspondences is essential~\cite{Labbe2019RTAB-Map,Mur2017ORB2,schoenberger2016sfm,schoenberger2016mvs}. The recent success of deep learning in various computer vision applications has motivated researchers to work on learned CNN-based approaches for interest point detection~\cite{Barroso2019KeyNet,Lenc2019Detector,Mishkin2018AffNet,Savinov2017detector,Verdie2015TILDE,zhang2017detector}, description~\cite{Balntas2016Localdesc,He2018LocDesc,Luo2019Contextdesc,Mishchuk2017HardNet,Tian2007L2Net,Tian2019SOSNET} or both detection and description~\cite{Detone18superpoint,Dusmanu2019D2Net,Germain2020S2DNet,Luo2020ASLFeat,Ono2018LFNet,Revaud2019R2D2,wang2021p2net,Yi2016LIFT}. Although, those approaches have demonstrated improved performance over classical hand-crafted methods on challenging benchmarks such as vision localization and image matching, most of them require ground-truth pixel correspondences between two views~\cite{Dusmanu2019D2Net,Jiang2021COTR,Revaud2019R2D2,Sun2021LoFTR}. Such accurate ground-truth labels are quite tedious to obtain since it requires information about the 3D representation of the scene (\eg 3D reconstruction, dense depth maps or scene flow). To address this issue, recent research directions have shifted towards weakly- and self-supervised methods. These methods rely on the supervision from epipolar geometry~\cite{Wang2020CAPS,Yang2019EPIPOLAR,Heng2019MONET,Zhou2021Patch2Pix}, principles of Reinforcement Learning~\cite{Bhowmik2020ReinforcedPoints,Truong2019GLAMpoints,Tyszkiewicz2020DISK}, relative camera poses~\cite{Bhowmik2020ReinforcedPoints,Wang2020CAPS}, camera intrinsics~\cite{Heng2021GeomPerc}, and rigid synthetic 2D transformations~\cite{Christiansen2019Unsuperpoint,Detone18superpoint,Revaud2019R2D2,tang2020NeuralRejection}. In this work, we show that with careful choices regarding hard-negative mining, color augmentation and photorealistic image stylization, we can achieve superior performance to supervised methods using only synthetic homographies. In principle, the proposed ideas can be also applied to supervised CNN-based approaches.

\customparagraph{Illumination invariance in feature descriptors.} Learning features that are invariant to illumination changes is a challenging task. To increase robustness to illumination changes, methods like~\cite{Dusmanu2019D2Net,Wang2020CAPS,Heng2021GeomPerc,Yang2020UR2KID} utilize a large SfM database of images with different lighting conditions. Domain adaptation-based methods~\cite{Anoosheh2020S2Night2Day,Mueller2019Image2Image,Porav2018Gans} utilize GANs~\cite{Goodfellow2014GANs,Zhu2017CycleGAN}, view synthesis~\cite{Mueller2019Image2Image} and few-shot learning~\cite{Baik2020DomainAdaptation} to reduce the domain gap between the training and the test (target) image distributions. Leveraging photorealistic image stylization to improve robustness to day-night variations was introduced in~\cite{Revaud2019R2D2}. In the follow-up work, Melekhov~\etal~\cite{Melekhov2020Stylization} extend the original method~\cite{Revaud2019R2D2} by considering more diverse style categories. Pautrat~\etal~\cite{Pautrat2020LISRD} propose a model predicting several dense descriptors with different levels of invariance. For a particular input image, the most relevant invariance is automatically selected using the concept of meta-descriptors based on a NetVLAD~\cite{Arandjelovic2016NetVLAD} layer. Inspired by~\cite{Melekhov2020Stylization,Revaud2019R2D2}, we utilize the idea of image stylization to improve the robustness of local descriptors to appearance variations.

\customparagraph{Self-supervised learning.} Self-supervised learning paves the way for training models on large amount of unlabelled data. Recent methods~\cite{chen2020SimCLR,he2020moco} in the classification domain are based on contrastive loss where representations from augmented copies of the original image -- positives, are contrasted against representations from other images in the batch or memory bank -- negatives. The quality of negatives was analyzed by several methods~\cite{debiased,kalantidis,robinson_2021} in improving self-supervised image representation learning. For local descriptors,~\cite{Laskar2020DensePixelMatching,dgcnet,rocco_17} consider only the augmented input image pair to sample positive and negative local descriptors (in-pair sampling). In this work, we propose methods to improve the quality of both positives and negatives. %

%% file: sections/method.tex
\section{Method}\label{sec:method}
In this section, we describe our method for local image descriptors learning that takes a single RGB image and produces a set of local feature descriptors in a self-supervised manner. Supervised methods leverage known scene geometry and utilize image- and keypoint level correspondences to increase robustness and invariance of local features to viewpoint and appearance variations. However, obtaining such ground-truth correspondence might be difficult especially for large-scale scenes. %
We analyze and carefully tailor improvements to existing self-supervised methods for learning invariant local descriptors.

We first review the main components of the proposed approach and the key ideas behind it, then we describe the hard-negative sampling procedure and training loss. An overview of our method is schematically illustrated in Figure~\ref{fig:pipeline_overview}. Given an input image $I_i \in \mathbb{R}^{H\times W\times 3}$ and a set of interest points $\mathbf{k}_i$, our goal is to predict local descriptors $\mathbf{d}_i$ corresponding to index $i$ which would be robust to appearance changes and have good generalization performance. To obtain interest points, we use SuperPoint~\cite{Detone18superpoint} detector. 

\customparagraph{Illumination Invariance}
Illumination invariance can be seen as learning context independent representations. In classification domain, various forms of CA methods such as RandAugment~\cite{Cubuk2019RandAugment} aim to remove the contextual bias in image representations. However, these methods apply various illumination transformations uniformly over both salient foreground object and background context which might be troublesome for local descriptor learning. %
Following~\cite{Melekhov2020Stylization,Revaud2019R2D2} we rely on image stylization~\cite{Li2018Stylization} to improve the robustness of local features under drastic illumination variations. Li~\etal~\cite{Li2018Stylization} propose a closed-form solution to image stylization based on a deep learning model trained in a self-supervised fashion by minimizing the sum of the reconstruction loss and perceptual loss. The method photorealistically transfers style of a reference photo to a content photo preserving local scene geometry. We use this pre-trained model as a part of our pipeline (\textbf{Style CNN} in Figure~\ref{fig:pipeline_overview}) and manually select reference images representing two style categories, \ie night and dusk, from the contributed views of the Amos Patches dataset~\cite{Pultar2019Amos}. In contrast to~\cite{Revaud2019R2D2}, for each category we consider \textit{multiple} style images which are then randomly applied to each image in the training set. The style categories and image stylization results are provided in the supplementary material.

\customparagraph{Viewpoint Invariance.} 
To model viewpoint invariance, recent self-supervised methods~\cite{Christiansen2019Unsuperpoint,Detone18superpoint}
apply a known homography transformation~\textbf{H},~\ie $\hat{I}=\mathbf{H}\left( I \right)$ followed by random crop. As in~\cite{Detone18superpoint}, we use similar types of the homographic adaptation operations: translation, rotation, perspective transform, and scale whose parameters are sampled from a uniform distribution during training (\textbf{H-crops generator}). In addition to stylization, we apply synthetic CA independently on the anchor and positive crops by adding blur, random per-pixel Gaussian noise, adaptive histogram equalization (CLAHE) along with augmentation in brightness, contrast, hue, and saturation. The full list of augmentations is also presented in supplementary.

\customparagraph{Descriptor CNN}. We use a fully-convolutional neural network architecture which takes an input crop $C_i\in \mathbb{R}^{H_c\times W_c\times 3}$ and outputs a dense map of L2-normalized fixed length descriptors, \ie $\mathbb{R}^{H_c\times W_c \times d}$. In this work, we consider two strong baseline methods (supervised and weakly-supervised), namely R2D2~\cite{Revaud2019R2D2} and CAPS~\cite{Wang2020CAPS}. Sec.~\ref{ssec:baselines} provides an overview of these baselines in more detail. In order to perform a fair comparison, we directly use R2D2 and CAPS models as a structure of Descriptor CNN.

\customparagraph{Hard-negative (HN) mining block.}
Supervised metric learning requires both positive and negative samples to learn meaningful representations. Without both positives and negatives, the model will collapse by converging to 0-valued representations. %
Existing unsupervised approaches~\cite{Detone18superpoint,Revaud2019R2D2} for learning local descriptors sample positives and negatives from a homography related image pair $\left(I, \hat{I}\right)$ with local descriptors $A \in I, P \in \hat{I}$ respectively. The images in the pair are dubbed as the anchor ($I$) and to the positive ($\hat{I}$). Each local descriptor $a_i \in A$ is associated with one $p_i \in P$ due to known homography matrix $\mathbf{H}$. Next, for each matching pair $\left(a_i, p_i\right)$ the index $n$ of the non-matching descriptor $p_n$ is obtained as follows:

\begin{equation}
    n = \argmax_{n=1..M, n\neq i} s\left(a_i, p_n\right),
\end{equation}

\noindent where $M$ is the number of interest points in $I$; $s(.)$ is a similarity function such as a dot-product. We refer to this baseline of selecting negatives as in-pair sampling. Several works~\cite{Gordo2017Retrieval,Mishchuk2017HardNet,Radenovic2016GEM} have shown that the quality of negatives has a strong impact in supervised learning of robust representations. Inspired by~\cite{Mishchuk2017HardNet}, we propose in-batch hard negatives sampling strategy that searches non-matching descriptors over all interest point embeddings $p \in B \setminus I$, where $B \sim D$ is a mini-batch sampled from the dataset $D$.

Finding the local hard negatives is an exhaustive process. Sampling negatives over a search space $D_n \sim D$ which is much larger than $B,\quad |D_n|\gg|B|$ ideally should improve the quality of negatives but incurs significant computational cost. Rather than mining hard-negatives at the local descriptor level, we propose a coarse-to-fine search method (\textbf{Image retrieval block} in Figure~\ref{fig:pipeline_overview}). For each image $I \in B$ we first find its nearest neighbor $I^n \in D_n$ in the feature space using visual descriptor $g_n$. This results in a mini-batch $B_n$ consisting of hard-negative image samples (the images look similar to the anchors $I \in B$ but \textit{not} homography related as the positives), \ie, $B_n=\{I_i^n\}_{i=1..|B|}$, which are then used to mine non-matching feature descriptors at the local level. Given the batch, $B_n$ the local descriptor pool now consists of $p_n \in B^\prime \setminus I$ where $B^\prime = (B \cup B_n)$. Such coarse-to-fine strategy leveraging visual image descriptors makes the process of mining hard-negative local descriptors tractable. Computing visual descriptors can be formulated in two ways. On the one hand, one can use state of the art image retrieval models, such as~\cite{Radenovic2016GEM,Radenovic2018TPAMI}, trained on supervised data generated by SfM systems to extract global descriptor from the entire image. On the other hand we propose a completely self-supervised approach whereby the global descriptors are constructed from the set of local descriptors extracted using the current network state. These local descriptors are then aggregated by some mean operation, such as GeM~\cite{Radenovic2016GEM}. %
Specifically, given a dense feature map $X_f \in \mathbb{R}^{H_c\times W_c \times N_d}$, visual descriptor $g_f \in \mathbb{R}^{N_d}$ is obtained by summing \textit{l2} normalized $X_f$ over the spatial resolution. The final representation is again \textit{l2} normalized.

\customparagraph{Loss functions.} Similar to~\cite{He2018LocDesc,Revaud2019R2D2}, we formulate our problem as the optimization of a differentiable approximation ($AP^\prime$) of the Average Precision (AP) ranking metric at training time:

\begin{equation}\label{eq:ap_loss}
    \mathcal{L}_{AP} = \frac{1}{N_a}\sum_i\left(1 - AP^\prime\left(a_i, p_i, \{p_{n_k}^i\}_{k=1..K}\right) \right)
\end{equation}
\noindent where $N_a$ is the number of anchor descriptors in a mini-batch $B^\prime$. In contrast to image matching~\cite{He2018LocDesc}, the number of positive and negative samples is highly imbalanced in our setting. Indeed, for each anchor descriptor in $B^\prime$ there is only one positive sample and several hundred negatives. To address this issue, we propose to select $top$-$K$ hard-negative local descriptors $\{p_{n_k}^i\}_{k=1..K}$ for $a_i$. Results presented in Sec.~\ref{ssec:results} demonstrate consistent improvement over the original AP loss~\cite{He2018LocDesc,Revaud2019R2D2}. For Triplet loss, the proposed idea of intra-batch hard-hegative mining also demonstrates better performance compared to in-pair sampling (see supplementary for more details).

%% file: sections/experiments.tex
\section{Experiments} \label{sec:experiments_results}
Here, we describe the experimental setting and validate that (1) increasing the search space from within-pair to intra-batch for hard negative mining improves the discriminativeness of local image descriptors and leads to better performance, (2) photorealistic image stylization in addition to color augmentations demonstrates good generalization performance and improves results, especially when query images are taken during challenging illumination conditions, (3) self-supervised methods for local image descriptors learning achieve competitive results with their fully- and weakly-supervised counterparts. In order to highlight the effectiveness of the proposed ideas, we further present an extensive ablation study.

\input{assets/plots/plt_mma}

\subsection{Baselines}\label{ssec:baselines}
In this work, we consider two recent approaches, R2D2~\cite{Revaud2019R2D2} and CAPS~\cite{Wang2020CAPS}, as the baseline methods. Since we focus exclusively on local image descriptors learning, SuperPoint~\cite{Detone18superpoint} keypoint detector is used in all our experiments and the baselines.

\customparagraph{R2D2}~\cite{Revaud2019R2D2} can directly extract dense descriptors and a heatmap of keypoints from an input image. It consists of L2-Net backbone network with two heads for jointly learning repeatable and reliable matches. The network was trained on pairs of image crops with known pixel correspondences based on either synthetic homography or 3D reprojection. We consider the model trained with 3D supervision as our baseline.

\customparagraph{CAPS}~\cite{Wang2020CAPS} is a CNN-based local image descriptor trained in a weakly-supervised manner. Specifically, it can learn feature descriptors using only scene labels and relative camera poses between views. It utilizes SuperPoint keypoints and an epipolar constraint on pixel locations of matched points as a supervision signal. The resulting descriptor consists of two concatenated 128-dimensional feature vectors, \ie coarse- and fine-level descriptors, extracted at different parts of the model encoder (an ImageNet-pretrained ResNet-50 architecture~\cite{Deng2009ImageNet,He2016ResNet}). In our experiments, we utilize \textit{only} fine-level descriptors to optimize the proposed criterion (Eq.~\ref{eq:ap_loss}).

\input{assets/images/fig_qualitative_matches}

\input{assets/tables/res_localization_all}

\subsection{Notations}\label{ssec:notations}
We use the following symbolic notation throughout the paper: {$\left(\squarehvfill\right)$} denotes models trained with synthetic CA; {$\left(\squarerightblack\right)$} indicates photorealistic stylized images; {$\left(\squarehvfill,\squarerightblack\right)$} if both CA and style transfer are used during training. Finally, we refer the models utilizing image retrieval pre-trained networks for extracting visual descriptors with a suffix~\texttt{*-gd} and~\texttt{*-selfgd} denotes the self-supervised way of computing visual descriptors on the fly using the current state of the proposed Descriptor CNN (\cf Figure~\ref{fig:pipeline_overview}).

\subsection{Benchmarks and metrics}\label{ssec:benchmarks_metrics}
We verify the proposed ideas on the following benchmarks.

\customparagraph{Sparse feature matching.} For this experiment, we evaluate our descriptors on the full images provided by the HPatches dataset~\cite{Balntas2017HPatches}. The dataset consists of 116 image sequences with varying photometric and viewpoint changes. Each sequence contains a reference image and 5 source images taken under a different viewpoint. For all image pairs, the estimated ground-truth homography matrix with respect to the reference image is provided. We follow the standard evaluation protocol~\cite{Dusmanu2019D2Net} and use the mean matching accuracy (MMA) as the metric, \ie the average percentage of correct matches per image under a certain pixel threshold. In addition to MMA, we also report a homography estimation score~\cite{Detone18superpoint,Dusmanu2019D2Net}, precision and recall computed for a 3-pixel error threshold. %

\customparagraph{Image-based localization.} To verify the generalization performance of the proposed approach, we evaluate our method on indoor and outdoor visual localization benchmarks. Specifically, we consider the large-scale \textit{outdoor} Aachen Day-Night v1.1~\cite{Zhang2020AachenV11} and Tokyo24/7~\cite{Torii2015Tokyo247} datasets for localization under severe illumination conditions. Image-based localization in complex \textit{indoor} environments is challenging due to the large viewpoint changes and weakly textured scenes. We use InLoc~\cite{Taira2018InLoc}, a large-scale indoor dataset with strong appearance changes between query images and the reference 3D map. The localization performance on the Aachen Day-Night v1.1 and InLoc benchmarks is reported as the percentage of correctly localized queries. For InLoc, we follow the localization pipeline\footnote{\url{https://github.com/cvg/Hierarchical-Localization}} proposed in~\cite{sarlin2019coarse, sarlin2020superglue}. The query is successfully localized if its camera position and orientation are both within a certain threshold.

The performance on the Tokyo24/7 dataset is evaluated using Recall@N, which is the number of queries that are correctly localized given $N$ nearest neighbor database images. Specifically, for a given query we first obtain a ranked list of database images, $L$ based on Euclidean distance between their global NetVLAD representations. The top-100 ranked database
images, $L^{\prime} \in L$ are re-ranked according to their similarity score based on the number of geometrically verified inliers between the query and each database image.

\customparagraph{Image retrieval.} We evaluate the proposed local image descriptors on the image retrieval benchmark. Specifically, the revisited Oxford ($\cR$Oxford5k) and revisited Paris ($\cR$Paris6k) datasets proposed by Radenovi{\'c}~\etal~\cite{Radenovic2018Revisited} have been used. Similarly to the Tokyo24/7 evaluation protocol, for each query image we first create a ranked list of database images based on cosine similarity of their global representations in the feature space and then compute
2D-2D correspondences between query and top-100 database images which are then re-ranked based on the number of inliers verified by RANSAC. To compute global descriptors, the model "retrievalSfM120k-resnet101-gem" proposed in~\cite{Radenovic2016GEM} is used. We evaluate performance by mean average precision (mAP) and mean precision at 1, 5, and 10 (mP@k), as defined by the protocol~\cite{Radenovic2018Revisited}.

\input{assets/tables/res_retrieval}

\subsection{Results}\label{ssec:results}
First, we experimentally compare the proposed models with their supervised oracles~\cite{Revaud2019R2D2,Wang2020CAPS} on the \textbf{sparse feature matching} benchmark by calculating MMA for different pixel thresholds illustrated in Figure~\ref{fig:hpatches_performance}. Following D2-Net~\cite{Dusmanu2019D2Net}, we separately report results for scenes corresponding to illumination changes (Illumination), viewpoint changes (Viewpoint), and for the whole dataset (Overall). The proposed unsupervised version of R2D2 (R2D2-U) shows a clear improvement over its supervised counterpart (R2D2-S) achieving better results for all evaluation thresholds. At the same time, although CAPS-U demonstrates competitive overall performance, it falls slightly behind its supervised baseline (CAPS-S~\cite{Wang2020CAPS}) on the illumination benchmark at larger pixel thresholds ($>$3px). Our hypothesis is that the original CAPS~\cite{Wang2020CAPS} model that leverages coarse-to-fine architecture (\cf Sec.~\ref{ssec:benchmarks_metrics}) allows to learn more discriminative descriptors leading to better matching performance. We also report homography estimation accuracy, precision (the percentage of correct matches over all the predicted matches) and recall (the ratio of correct matches over the total number of ground truth correspondences) using the same HPatches dataset. Following~\cite{Pautrat2020LISRD}, we use mutual nearest neighbor matcher and a threshold of 3 pixels to consider a match to be correct. The results are summarized in Table~\ref{tbl:hpatches_eval}. As can be seen, the proposed unsupervised models demonstrate competitive performance. Interestingly, the model utilizing global image descriptor for hard-negative mining (CAPS-{$\left(\squarehvfill,\squarerightblack\right)$}-gd) achieves the best Recall and Precision performance over all other methods.

\customparagraph{Visual localization.} We evaluate the performance of the proposed pipeline on the task of image-based localization. Note, that we train our approaches only on outdoor datasets,~\ie Megadepth and Phototourism but evaluate on both outdoor (Aachen v1.1) and indoor (InLoc) to test the generalization of the proposed models. We use R2D2, CAPS and R2D2* as baseline models where R2D2* represents R2D2 model trained with SuperPoint detector, \ie, the setup which is closer to our pipeline (see Sec.~\ref{sec:method}).  For unsupervised methods, we report results obtained by the models utilize in-batch hard-negative sampling during training. The percentage of correctly localized query images is presented in Table~\ref{tbl:loc_aachen_inloc}. First, if we compare R2D2-{$\left(\squarehvfill\right)$} and R2D2-{$\left(\squarehvfill,\squarerightblack\right)$} trained on Aachen database images (A), we can notice that using only synthetic color augmentation is not sufficient to handle drastic illumination changes. Increasing the number of training samples leads to better localization performance. The model R2D2-{$\left(\squarehvfill,\squarerightblack\right)$} trained on (M+P) is comparable to the supervised counterpart on day-time and surpasses it on night-time queries by a noticeable margin: up to $+\textbf{0.5}$ for $\left(0.5m, 5^\circ\right)$ and $+\textbf{1.1}$ for $\left(5m, 10^\circ\right)$. For indoor pose estimation, the same model can outperform the supervised baseline in all metrics: up to $+\textbf{1.5}$ and $+\textbf{2.9}$ for the finest threshold $\left(0.25m, 2^\circ\right)$ for scene DUC1 and DUC2 respectively. Leveraging visual descriptors (*-gd and *-selfgd) provides better outdoor localization results over the models trained with only stylization and color augmentations (R2D2-{$\left(\squarehvfill,\squarerightblack\right)$} and CAPS-{$\left(\squarehvfill,\squarerightblack\right)$}). However, in indoors, visual descriptors \textit{cannot} provide strong generalization performance leading to a marginal improvement (for the R2D2 backbone) or even poor results (with the CAPS backbone). We hypothesize that such behaviour is caused by larger gap in data distribution during training and evaluation. Interestingly, the transferring accuracy depends on the Descriptor CNN (\cf Figure~\ref{fig:pipeline_overview}) structure. The unsupervised model trained on (M+P) and based on the CAPS backbone \textit{cannot} improve performance of outdoor localization compared to its supervised baseline (CAPS~\cite{Wang2020CAPS}). However, large gains are seen for indoor setting: up to $+\textbf{6.1}$ (34.4 \textit{vs.} 40.5) for the most rigorous threshold of scene DUC2. Table~\ref{tbl:loc_tokyo247} shows the localization performance on Tokyo24/7. Similarly to the results on Aachen, R2D2-{$\left(\squarehvfill,\squarerightblack\right)$} is \textit{better} than its supervised counterpart (R2D2-[A+R]) in all metrics: $+\textbf{1.58}$, $+\textbf{1.9}$, and  $+\textbf{1.9}$, respectively. Qualitative results of the proposed models on test images of MegaDepth are illustrated in Figure~\ref{fig:qualitative_results}.

\customparagraph{Image retrieval.} As shown in Table~\ref{tbl:image_retrieval_results}, the proposed unsupervised models show competitive image retrieval performance and surpass their supervised counterparts in mP@k and mAP on $\cR$Oxford5k and $\cR$Paris6k. Most notably, utilizing visual image descriptors with the model based on CAPS backbone leads to better retrieval accuracy which is consistent with the results obtained for the camera localizaton benchmark. However, for the R2D2 backbone model, mining harder negative samples with the help of visual descriptors leads to a marginal improvement (R2D2-{$\left(\squarehvfill,\squarerightblack\right)$} \textit{vs.} R2D2-{$\left(\squarehvfill,\squarerightblack\right)$}-gd) which can be explained by the network capacity. In $\cR$Paris6k both the supervised and unsupervised methods attain comparable performance. Notably, the biggest performance improvement of 5\% is observed across mP@10 under Hard setting.

We observe that increasing the size of dataset ($A \rightarrow M \rightarrow M+P$) performs favourably on the different benchmarks. Stylization provides substantial improvements on Aachen Night time dataset, while performs comparably on other benchmarks. This indicates that the proposed network is not biased to the chosen styles. Furthermore, the proposed coarse-to-fine method with global descriptors also performs favourably compared to in-batch sampling especially on Aachen and $\cR$Oxford5k datasets. Finally, the self-supervised global mining approach *-selfgd performs comparably to *-gd which utilizes image retrieval models pre-trained on SfM data. Improving the on the fly strategy of mining hard negative local samples conditioned on visual image descriptors is a possible future direction.  %

%% file: assets/plots/plt_mma.tex
\begin{figure*}[t!]
    \begin{minipage}{0.65\linewidth}
    \centering
    \includegraphics[width=\textwidth]{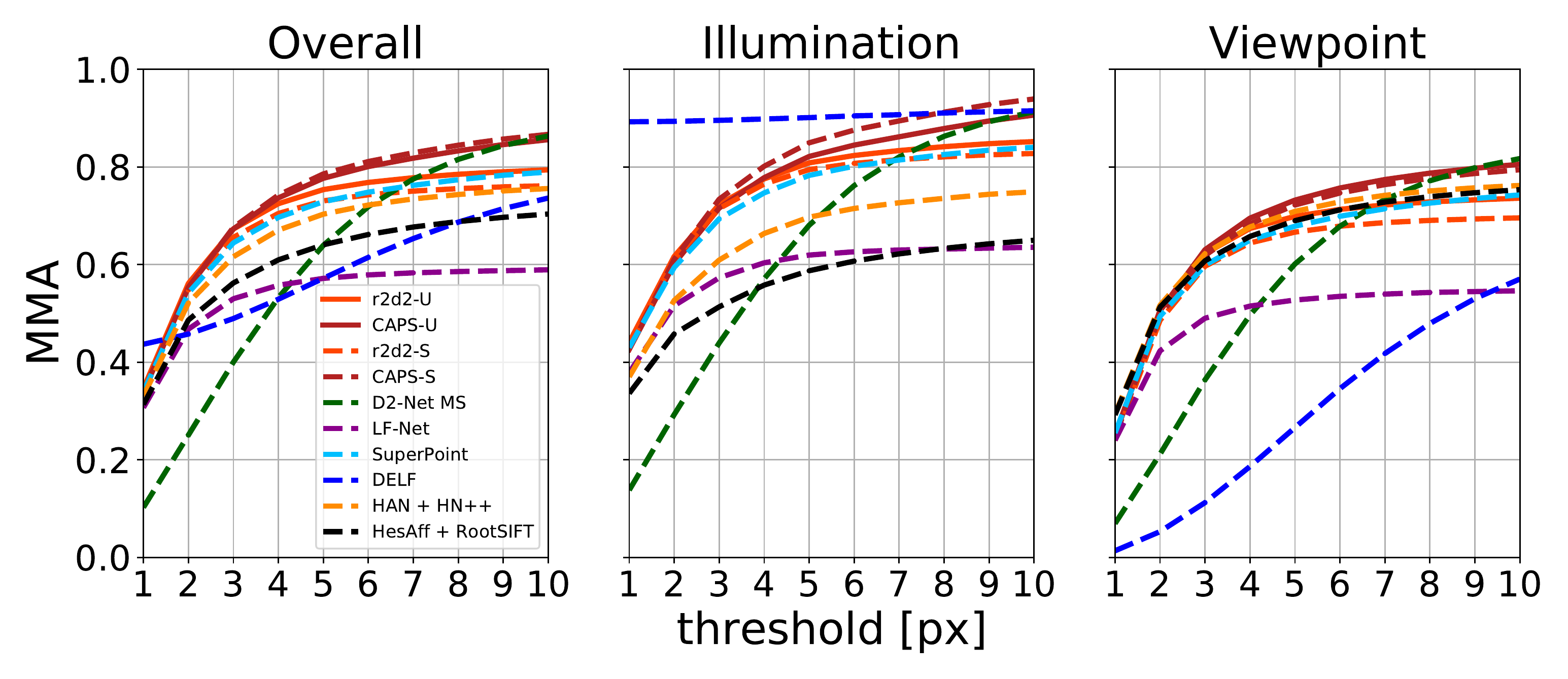}
    \caption{\textbf{Mean matching accuracy (MMA) on the HPatches~\cite{Balntas2017HPatches} dataset.} For each method, the MMA with different pixel error thresholds is reported. The proposed self-supervised models (R2D2-U and CAPS-U) %
    achieve better overall performance for stricter matching thresholds (up to 3 pixels) compared to their supervised (R2D2-S and CAPS-S) counterparts. To avoid image clutter, for R2D2-U and CAPS-U, we only provide the results obtained by the {$\left(\squarehvfill,\squarerightblack\right)$} models.}
    \label{fig:hpatches_performance}
    \end{minipage}
    ~
    \begin{minipage}{0.34\linewidth}
        \centering
        \resizebox{\textwidth}{!}{%
		\footnotesize
		\begin{tabular}{l | c c c | c c c }
		\toprule
		\multirowcell{2}{Method} & \multicolumn{3}{c|}{Illumination} & \multicolumn{3}{c}{Viewpoint} \\
		& \textEta & Precision & Recall & \textEta & Precision & Recall \\
		\midrule\midrule
		Root SIFT~\cite{Arandjelovic2012RootSIFT} & 0.933 & \textbf{0.782} & 0.799 & 0.566 & 0.651 & 0.527\\
		HardNet~\cite{Mishchuk2017HardNet} & 0.940 & 0.702 & 0.731 & 0.664 & 0.701 & 0.734\\
		SOSNet~\cite{Tian2019SOSNET} & 0.933 & 0.748 & 0.821 & \textbf{0.698} & 0.727 & 0.760\\
		SuperPoint~\cite{Detone18superpoint} & 0.912 & 0.710 & 0.811 & 0.671 & 0.685 & 0.750\\
		D2-Net~\cite{Dusmanu2019D2Net} & 0.905 & 0.725 & 0.775 & 0.617 & 0.666 & 0.664\\
		LISRD~\cite{Pautrat2020LISRD} & \textbf{0.947} & 0.766 & 0.920 & 0.688 & 0.731 & 0.757\\
		\textcolor{Magenta}{R2D2}~\cite{Revaud2019R2D2} & 0.940 & 0.762 & 0.837 & 0.692 & 0.720 & 0.732\\
		\textcolor{Cyan}{CAPS}~\cite{Wang2020CAPS} & 0.888 & 0.757 & \textcolor{Cyan}{\underline{\textbf{0.938}}} & \textcolor{Cyan}{\underline{0.692}} & 0.723 & 0.699\\
		\midrule\midrule
		R2D2-{$\left(\squarehvfill,\squarerightblack\right)$} & 0.944 & 0.764 & \textcolor{Magenta}{\underline{0.838}} & 0.678 & \textcolor{Magenta}{\underline{0.732}} & \textcolor{Magenta}{\underline{0.739}} \\
		R2D2-{$\left(\squarehvfill,\squarerightblack\right)$}-selfgd & 0.933 &	0.761 &	0.817 &	0.678 &	0.715 &	0.705  \\
		R2D2-{$\left(\squarehvfill,\squarerightblack\right)$}-gd & \textcolor{Magenta}{\underline{\textbf{0.947}}} &	\textcolor{Magenta}{\underline{0.766}}	& 0.826 &	\textcolor{Magenta}{\underline{\textbf{0.698}}} &	0.726 &	0.720 \\
		CAPS-{$\left(\squarehvfill,\squarerightblack\right)$} & 0.933 & 0.750 & 0.884 & 0.671 & 0.742 & 0.728 \\
		CAPS-{$\left(\squarehvfill,\squarerightblack\right)$}-selfgd & \textcolor{Cyan}{\underline{0.937}} &	0.756 &	0.893 &	0.661 &	0.740 &	0.752\\
		CAPS-{$\left(\squarehvfill,\squarerightblack\right)$}-gd & 0.919 & \textcolor{Cyan}{\underline{0.757}} & 0.890 & 0.681 & \textcolor{Cyan}{\underline{\textbf{0.747}}} & \textcolor{Cyan}{\underline{\textbf{0.762}}}\\
		\end{tabular}}
		\captionof{table}{\textbf{Evaluation results on HPatches.} We follow~\cite{Pautrat2020LISRD} and report the accuracy of homography estimation, precision and recall for error thresholds of 3 pixels. The best score between supervised baselines, \textcolor{Magenta}{R2D2} and \textcolor{Cyan}{CAPS}, and the proposed methods for each category is \underline{underlined}. The overall best score is in \textbf{bold}.}\label{tbl:hpatches_eval}
    \end{minipage}
    
\end{figure*}

%% file: assets/images/fig_qualitative_matches.tex
\begin{figure*}[t!]
  \centering
    \includegraphics[width=1\linewidth]{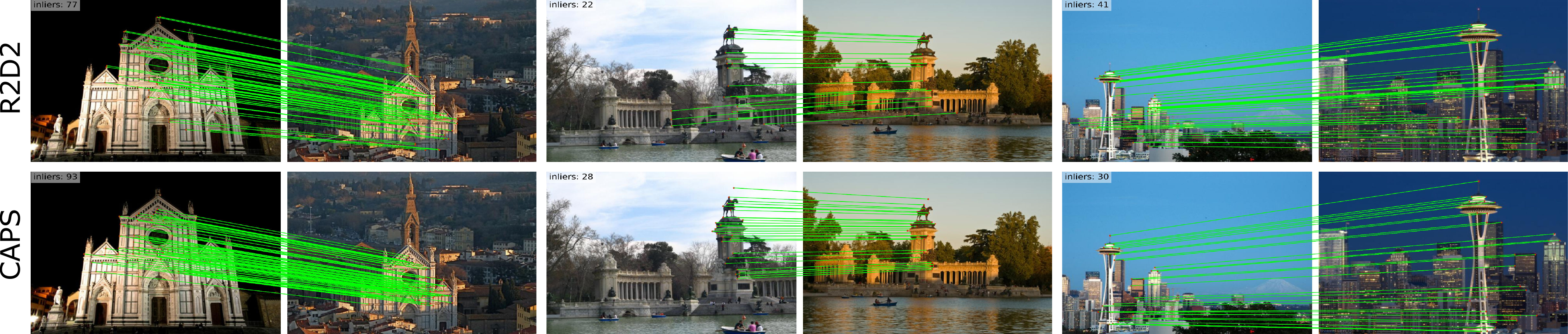}
 \caption{\textbf{Qualitative matching results.} We evaluate the proposed unsupervised pipeline with two different backbone models, R2D2 (top row) and CAPS (bottom row) on test samples of the MegaDepth dataset~\cite{Li2018Megadepth}. Both models have been trained on stylized images with color augmentations using in-batch hard negative sampling. Green lines indicate correspondences estimated by RANSAC (best viewed digitally). The proposed method predicts reliable matches even without training on annotated correspondences.}
 \label{fig:qualitative_results}
\end{figure*}

%% file: assets/tables/res_localization_all.tex
\begin{table*}[t!]
    \begin{subtable}[h]{0.7\textwidth}
        \begin{center}
	    \resizebox{\textwidth}{!}{%
		\footnotesize 		
		\begin{tabular}{c l | c c | c c c | c c c | c c c | c c c}
		\toprule
		\multirow{4}{*}{} & \multirowcell{4}{Method} & \multirowcell{4}{Supervision} & \multirowcell{4}{Training\\data} & \multicolumn{6}{c}{Aachen v1.1} & \multicolumn{6}{|c}{InLoc} \\
		& & & & \multicolumn{6}{c}{\% localized queries} & \multicolumn{6}{|c}{\% localized queries} \\
		& & & & \multicolumn{3}{c|}{Day (824 images)} & \multicolumn{3}{c}{Night (191 images)} & \multicolumn{3}{|c|}{DUC1} &  \multicolumn{3}{c}{DUC2} \\
		& & & & $0.25m, 2^\circ$ & $0.5m, 5^\circ$ & $5m, 10^\circ$ & $0.25m, 2^\circ$ & $0.5m, 5^\circ$ & $5m, 10^\circ$ & $0.25m, 2^\circ$ & $0.5m, 5^\circ$ & $5m, 10^\circ$ & $0.25m, 2^\circ$ & $0.5m, 5^\circ$ & $5m, 10^\circ$ \\ \midrule\midrule
        \multirow{3}{*}{\rotatebox[origin=c]{90}{Super}} & \textcolor{Magenta}{R2D2}~\cite{Revaud2019R2D2} & OF & A+R & \textcolor{Magenta}{\textbf{88.6}} & \textcolor{Magenta}{\textbf{95.4}} & \textcolor{Magenta}{\textbf{98.9}} & 72.8 & 89.0 & 97.4 & 27.8 & 42.4 & 54.5 & 22.1 & 34.4 & 42.7 \\
        & \textcolor{Magenta}{R2D2*} & OF & A & 87.7 & 94.7 & 98.7 & 69.6 & 86.4 & 95.3 & 29.8 & 43.4 & 55.1 & 21.4 & 34.4 & 43.5 \\
        & \textcolor{Cyan}{CAPS}~\cite{Wang2020CAPS} & SL+RP & M & 85.3 & \textcolor{Cyan}{\textbf{93.8}} & 97.9 & \textcolor{Cyan}{\textbf{75.9}} & 88.5 & \textcolor{Cyan}{\textbf{97.9}} & 38.4 & 59.1 & \textcolor{Cyan}{\textbf{74.7}} & 34.4 & 52.7 & 61.8\\ \midrule\midrule
        \multirow{11}{*}{\rotatebox[origin=c]{90}{Self-supervised}} & R2D2-$\left(\squarehvfill\right)$ & - & A & 87.4 & 94.9 & 98.3 & 63.9 & 80.1 & 92.1 & \textcolor{Magenta}{\textbf{29.3}} & 46.5 & 57.6 & 29.8 & 42.0 & 51.1 \\
        & R2D2-{$\left(\squarehvfill,\squarerightblack\right)$} & - & A & 88.0 & 94.8 & 98.2 & 70.2 & 86.4 & 95.8 & 27.8 & 46.5 & 57.6 & 29.8 & 42.0 & 51.1\\
        & R2D2-{$\left(\squarehvfill,\squarerightblack\right)$} & - & M & 87.4 & 94.7 & 98.3 & 72.3 & 88.5 & 97.4 & 26.8 & 46.0 & 57.1 & 26.7 & \textcolor{Magenta}{\textbf{42.7}} & 48.9 \\
        & R2D2-{$\left(\squarehvfill,\squarerightblack\right)$}-gd & - & M & 87.5 & 94.9 & 98.3 & 71.7 & 86.4 & 96.9 & 27.8 & 43.4 & 54.5 & 28.2 & 41.2 & 51.9 \\
        & R2D2-{$\left(\squarehvfill,\squarerightblack\right)$}-selfgd & - & M & 88.1 & 94.8 & 98.1 & 71.2 & 88.0 & 95.8 & 30.8 & 43.9 & 54.0 & 28.2 & 37.4 & 46.6\\
        & R2D2-{$\left(\squarehvfill,\squarerightblack\right)$} & - & M+P & 88.2 & 95.1 & 98.5 & \textcolor{Magenta}{\textbf{73.3}} & \textcolor{Magenta}{\textbf{90.1}} & \textcolor{Magenta}{\textbf{97.4}} & \textcolor{Magenta}{\textbf{29.3}} & \textcolor{Magenta}{\textbf{47.0}} & \textcolor{Magenta}{\textbf{58.1}} & \textcolor{Magenta}{\textbf{25.2}} & 41.2 & \textcolor{Magenta}{\textbf{51.9}} \\ 
        \cmidrule{2-16}
        & CAPS-{$\left(\squarehvfill\right)$} & - & M & 85.8 & 93.8 & 98.2 & 67.0 & 82.2 & 96.9 &35.9 & 54.0 & 69.2 & 38.9 & 50.4 & 64.9\\
        & CAPS-{$\left(\squarehvfill,\squarerightblack\right)$} & - & M & 85.1 & 93.2 & 97.8 & 71.7 & 87.4 & \textcolor{Cyan}{\textbf{97.9}} & \textcolor{Cyan}{\textbf{39.9}} & \textcolor{Cyan}{\textbf{61.6}} & 70.7 & 38.2 & 51.1 & \textcolor{Cyan}{\textbf{66.4}}\\
        & CAPS-{$\left(\squarehvfill,\squarerightblack\right)$}-gd & - & M & \textcolor{Cyan}{\textbf{87.0}} & \textcolor{Cyan}{\textbf{93.8}} & \textcolor{Cyan}{\textbf{98.3}} & 73.8 & \textcolor{Cyan}{\textbf{89.0}} & 97.4 & 35.9 & 53.0 & 65.7 & 32.8 & 47.3 & 61.1\\
        & CAPS-{$\left(\squarehvfill,\squarerightblack\right)$}-selfgd & - & M & 86.9 & \textcolor{Cyan}{\textbf{93.8}} & 98.1 & 71.7 & \textcolor{Cyan}{\textbf{89.0}} & 97.4 & 35.9 & 53.5 & 64.6 & 30.5 & 45.0 & 58.8 \\
        & CAPS-{$\left(\squarehvfill,\squarerightblack\right)$} & - & M+P & 85.4 & 93.2 & 97.9 & 72.3 & 88.5 & \textcolor{Cyan}{\textbf{97.9}} & 38.9 & 56.1 & 69.2 & \textcolor{Cyan}{\textbf{40.5}} & \textcolor{Cyan}{\textbf{54.2}} & \textcolor{Cyan}{\textbf{66.4}}\\
		\end{tabular}
        }
	\end{center}
	\vspace{-5pt}
	\caption{Evaluations results for Aachen v1.1~\cite{Zhang2020AachenV11} and InLoc~\cite{Taira2018InLoc} }\label{tbl:loc_aachen_inloc}
    \end{subtable}
    ~
    \begin{subtable}[h]{.29\textwidth}
        	\begin{center}
	\resizebox{.9\textwidth}{!}{%
		\footnotesize
		\begin{tabular}{c l|l l l}
        \multirow{2}{*}{} & \multirow{2}{*}{Method} & \multicolumn{3}{|c}{Recall} \\
        & & \multicolumn{1}{|c}{r@1} & \multicolumn{1}{c}{r@5} & \multicolumn{1}{c}{r@10}\\
    \hline
    \hline
    & DenseVLAD~\cite{Torii2015Tokyo247} & 67.10 & 74.20 & 76.10 \\
    & NetVLAD-TokyoTM~\cite{Arandjelovic2016NetVLAD} & 71.10 & 83.10 & 86.20 \\
    & SIFT~\cite{Lowe2004SIFT} & 73.33 & 80.00 & 84.40 \\
    & InLoc~\cite{Taira2018InLoc} & 62.54 & 67.62  & 70.48 \\
    & Dense Pixel Matching~\cite{Laskar2020DensePixelMatching} & 77.14 & 84.44 & 86.67 \\
    & \textcolor{Magenta}{R2D2}-[A+R]~\cite{Revaud2019R2D2} & 78.10 & 82.54 & 84.13 \\
    & \textcolor{Cyan}{CAPS}-[M]~\cite{Wang2020CAPS} & \textcolor{Cyan}{\textbf{83.49}} & 86.67 & 87.94 \\ \hline \hline
    \multirow{9}{*}{\rotatebox[origin=c]{90}{Self-supervised}} & R2D2-{$\left(\squarehvfill,\squarerightblack\right)$}-[A] & 79.05 & 83.17 & 85.71 \\
    & R2D2-{$\left(\squarehvfill,\squarerightblack\right)$}-[M] & 79.37 & \textcolor{Magenta}{\textbf{84.76}} & 85.71 \\
    & R2D2-{$\left(\squarehvfill,\squarerightblack\right)$}-selfgd-[M] & 78.41 & 82.54 & 83.81 \\
    & R2D2-{$\left(\squarehvfill,\squarerightblack\right)$}-gd-[M] & 77.78 & 82.86 & 85.40 \\
    & R2D2-{$\left(\squarehvfill,\squarerightblack\right)$}-[M+P] & \textcolor{Magenta}{\textbf{79.68}} & 84.44 & \textcolor{Magenta}{\textbf{86.03}} \\ \cmidrule{2-5}
    & CAPS-{$\left(\squarehvfill,\squarerightblack\right)$}-[M] & 82.54 & 86.03 & 87.94 \\
    & CAPS-{$\left(\squarehvfill,\squarerightblack\right)$}-selfgd-[M] & 81.27 & 85.71 & 86.98 \\
    & CAPS-{$\left(\squarehvfill,\squarerightblack\right)$}-gd-[M] & 82.54 & 86.35 & 87.94 \\
    & CAPS-{$\left(\squarehvfill,\squarerightblack\right)$}-[M+P] & 82.86 & \textcolor{Cyan}{\textbf{87.30}} & \textcolor{Cyan}{\textbf{88.25}} \\

    \end{tabular}
		
	}
	
	\end{center}
	\vspace{-10pt}
	\caption{Performance on Tokyo24/7~\cite{Torii2015Tokyo247}}\label{tbl:loc_tokyo247}
    \end{subtable}
    
    \vspace{-5pt}
    \caption{\textbf{Indoor and outdoor localization performance}. The supervised baseline methods, \textcolor{Magenta}{R2D2} and \textcolor{Cyan}{CAPS}, are color-coded. Best results in each category for each localization benchmark are in \textbf{bold}. \textbf{Legend}: \textit{Supervision}: OF~--~ground-truth pixel correspondences based on optical flow~\cite{Revaud2019R2D2}; SL~--~ground-truth scene labels; RP~--~relative camera poses. \textit{Training data}: A~--~Aachen database images; R~--~Random images from the Internet~\cite{Revaud2019R2D2}; M~--~MegaDepth dataset; P~--~Phototourism dataset. For Tokyo24/7, the training datasets for our supervised baselines and the proposed approach are given in $\left[.\right]$ parentheses.}
    
\end{table*}

%% file: assets/tables/res_retrieval.tex
\begin{table*}[t!]
	\begin{center}
	\resizebox{.99\textwidth}{!}{%
		\footnotesize
		\begin{tabular}{c l | c c | c c c | c c c | c c c | c c c | c c c |  c c c | c c c | c c c}
		\toprule
		 \multirow{3}{*}{} & \multirowcell{3}{Method} & \multirowcell{3}{Training\\data} & \multirowcell{3}{Hard-negative\\mining} & \multicolumn{12}{|c}{$\cR$Oxford5k} & \multicolumn{12}{|c}{$\cR$Paris6k} \\
	     & & & & \multicolumn{3}{c}{mAP} & \multicolumn{9}{|c|}{mP@k [1, 5, 10]} & \multicolumn{3}{c}{mAP} & \multicolumn{9}{|c}{mP@k [1, 5, 10]} \\
		 & & & & E & M & H & \multicolumn{3}{c|}{E} & \multicolumn{3}{c|}{M} & \multicolumn{3}{c|}{H} & E & M & H & \multicolumn{3}{c|}{E} & \multicolumn{3}{|c|}{M} &  \multicolumn{3}{c}{H} \\ \midrule\midrule
        \multirow{4}{*}{\rotatebox[origin=c]{90}{Supervised}} & Baseline~\cite{Radenovic2016GEM} & - & - & 72.60 & 54.89 &	27.37 & 91.43 & 80.74 & 74.12 & 91.43 & 80.57 & 73.86 & 61.43 & 47.50 & 39.79 & 86.15 & 68.98 & 43.86 & 98.57 & 95.14 & 93.43 & 98.57 & 98.57 & 96.14 & 95.71 & 87.71 & 82.86\\
		& \textcolor{Magenta}{R2D2}~\cite{Revaud2019R2D2} & A+R & in-pair & 76.39 & \textcolor{Magenta}{\textbf{59.09}} & \textcolor{Magenta}{\textbf{33.20}} & 97.06 & 88.38 & 81.55 & 95.71 & 89.71 & 83.57 & \textcolor{Magenta}{\textbf{85.71}} & \textcolor{Magenta}{\textbf{61.14}} & \textcolor{Magenta}{\textbf{47.14}} & 85.73 & 68.51 & 42.86 & \textcolor{Magenta}{\textbf{100}} & 97.24 & 95.81 & \textcolor{Magenta}{\textbf{100}} & \textcolor{Magenta}{\textbf{100}} & 99.14 & \textcolor{Magenta}{\textbf{100}} & \textcolor{Magenta}{\textbf{92.57}} & 81.57\\
		& R2D2* & A & in-pair & 75.36 & 58.61 & 32.69 & 97.06 & 87.35 & 79.56 & 95.71 & 89.43 & 82.57 & 85.71 & 59.14 & 45.86 & 85.39 & 68.36 & 42.93 & 100 & 97.43 & 96.14 & 100 & 99.71 & 98.86 & 98.57 & 91.14 & 82.43\\
        & \textcolor{Cyan}{CAPS}~\cite{Wang2020CAPS} & M & - & 75.27 & 59.67 & \textcolor{Cyan}{\textbf{34.77}} & 94.12 & 86.76 & 80.44 & 94.29 & 89.14 & 84.14 & \textcolor{Cyan}{\textbf{85.71}} & \textcolor{Cyan}{\textbf{66.00}} & \textcolor{Cyan}{\textbf{50.57}} & \textcolor{Cyan}{\textbf{88.92}} & \textcolor{Cyan}{\textbf{69.87}} & \textcolor{Cyan}{\textbf{44.47}} & \textcolor{Cyan}{\textbf{100}} & \textcolor{Cyan}{\textbf{99.43}} & \textcolor{Cyan}{\textbf{97.43}} & \textcolor{Cyan}{\textbf{100}} & \textcolor{Cyan}{\textbf{100}} & \textcolor{Cyan}{\textbf{100}} & 98.57 & \textcolor{Cyan}{\textbf{94.86}} & \textcolor{Cyan}{\textbf{87.14}}\\ \midrule\midrule
		& R2D2-$\left(\squarehvfill\right)$ & M & in-pair & 70.99 & 55.38 & 29.67 & 95.59 & 84.12 & 75.44 & 94.29 & 86.57 & 78.43 & 82.86 & 54.57 & 42.29 & 84.57 & 67.95 & 42.57 & 100 & 97.14 & 96.00 & 100 & 100 & 99.14 & 100 & 90.29 & 81.86  \\
        \multirow{8}{*}{\rotatebox[origin=c]{90}{Self-supervised}} & R2D2-$\left(\squarehvfill\right)$ & A & in-batch & 73.16 & 56.47 & 30.05 & 95.59 & 85.51 & 77.87 & 94.29 & 87.14 & 79.29 & 82.86 & 55.43 & 42.57 & 86.09 & 68.60 & \textcolor{Magenta}{\textbf{43.04}} & 100 & 98.00 & 96.43 & 100 & 99.71 & 99.43 & 97.14 & 91.71 & \textcolor{Magenta}{\textbf{82.71}}  \\
		& R2D2-$\left(\squarehvfill\right)$ & M & in-batch & 75.09 & 58.69 & 33.17 & 97.06 & 87.06 & 80.15 & 95.71 & 89.71 & 83.29 & 85.71 & 60.29 & 45.71 & 86.38 & 68.60 & 42.83 & 100 & 98.00 & \textcolor{Magenta}{\textbf{96.86}} & 100 & 100 & \textcolor{Magenta}{\textbf{99.57}} & 98.57 & 91.43 & 82.57 \\
		& R2D2-{$\left(\squarehvfill,\squarerightblack\right)$} & M & in-batch & 75.70 & 58.79 & 33.03 & 97.06 & 86.57 & 80.88 & 95.71 & 89.14 & 82.57 & 88.57 & 60.00 & 46.29 & 86.49 & 68.67 & 42.86 & 100 & 98.29 & 96.43 & 100 & 100 & 99.29 & \textcolor{Magenta}{\textbf{100}} & 92.00 & 82.14 \\
		& R2D2-{$\left(\squarehvfill,\squarerightblack\right)$}-gd & M & in-batch & 75.97 & 58.70 & 32.85 & 97.06 & 88.24 & 81.03 & 95.71 & \textcolor{Magenta}{\textbf{90.00}} & 83.14 & 85.71 & 60.29 & 46.86 & 86.31 & 68.63 & 42.85 & 100 & 98.00 & 96.29 & 100 & 100 & 99.43 & 100 & 91.14 & \textcolor{Magenta}{\textbf{82.71}} \\
		& R2D2-{$\left(\squarehvfill,\squarerightblack\right)$}-selfgd & M & in-batch & 73.47 & 57.57 & 32.16 & 97.06 & 85.59 & 76.91 & 95.71 & 88.86 & 81.29 & 85.71 & 58.57 & 46.43 & 86.04 & 68.48 & 42.76 & 100 & 97.71 & 96.14 & 100 & 99.71 & 99.29 & 98.57 & 89.71 & 82.29 \\
		& R2D2-{$\left(\squarehvfill,\squarerightblack\right)$} & M+P & in-batch & \textcolor{Magenta}{\textbf{76.65}} & 59.07 & 33.15 & \textcolor{Magenta}{\textbf{97.06}} & \textcolor{Magenta}{\textbf{88.82}} & \textcolor{Magenta}{\textbf{81.62}} & \textcolor{Magenta}{\textbf{95.71}} & 89.43 & \textcolor{Magenta}{\textbf{83.71}} & \textcolor{Magenta}{\textbf{85.71}} & 60.57 & \textcolor{Magenta}{\textbf{47.14}} & \textcolor{Magenta}{\textbf{87.02}} & \textcolor{Magenta}{\textbf{68.80}} & 42.85 & \textcolor{Magenta}{\textbf{100}} & \textcolor{Magenta}{\textbf{98.57}} & \textcolor{Magenta}{\textbf{96.86}} & \textcolor{Magenta}{\textbf{100}} & \textcolor{Magenta}{\textbf{100}} & \textcolor{Magenta}{\textbf{99.57}} & 98.57 & 91.43 & 81.86 \\ \cmidrule{2-28}
		& CAPS-{$\left(\squarehvfill,\squarerightblack\right)$} & M & in-batch & 74.68 & 58.64 & 32.54 & 95.59 & 87.06 & 78.38 & 94.29 & 89.71 & 82.71 & 82.86 & 59.14 & 46.29 & 87.88 & 69.25 & 43.46 & 100 & 99.14 & 96.71 & 100 & 100 & 99.71 & 100 & 92.86 & 84.14 \\
		& CAPS-{$\left(\squarehvfill,\squarerightblack\right)$}-gd & M & in-batch & 77.06 & 59.67 & 32.89 & 95.59 & 88.82 & \textcolor{Cyan}{\textbf{82.04}} & 94.29 & 90.57 & \textcolor{Cyan}{\textbf{84.71}} & 82.86 & 60.00 & 47.57 & 87.36 & 69.19 & 43.54 & 100 & 98.86 & 96.71 &  100 & 100 & 99.86 & 100 & 94.00 & 84.71 \\
		& CAPS-{$\left(\squarehvfill,\squarerightblack\right)$}-selfgd & M & in-batch & \textcolor{Cyan}{\textbf{77.17}} & 59.54 & 32.71 & \textcolor{Cyan}{\textbf{97.06}} & \textcolor{Cyan}{\textbf{89.85}} & 81.47 & \textcolor{Cyan}{\textbf{95.71}} & \textcolor{Cyan}{\textbf{90.86}} & 84.43 & \textcolor{Cyan}{\textbf{85.71}} & 60.57 & 47.29 & 87.61 & 69.14 & 43.33 & \textcolor{Cyan}{\textbf{100}} & 98.86 & 96.71 & \textcolor{Cyan}{\textbf{100}} & \textcolor{Cyan}{\textbf{100}} & 99.57 & \textcolor{Cyan}{\textbf{100}} & 93.43 & 83.71\\
		& CAPS-{$\left(\squarehvfill,\squarerightblack\right)$} & M+P & in-batch & 76.49 & \textcolor{Cyan}{\textbf{59.72}} & 33.60 & 95.59 & 88.16 & 81.54 & 94.29 & 89.71 & 84.14 & 84.29 & 62.00 & 48.14 & 87.66 & 69.12 & 43.34 & \textcolor{Cyan}{\textbf{100}} & 98.86 & 97.00 & \textcolor{Cyan}{\textbf{100}} & \textcolor{Cyan}{\textbf{100}} & 100 & 100 & 93.14 & 84.71\\
		\end{tabular}
	}
	\caption{\textbf{Image retrieval performance}. We report results for the $\cR$Oxford5k and $\cR$Paris6k datasets following the evaluation protocol proposed in~\cite{Radenovic2018Revisited}. We use the same color-map for the supervised baseline approaches as in Table~\ref{tbl:loc_aachen_inloc}. The proposed unsupervised models demonstrate competitive performance 
	\vspace{-5mm}
	\label{tbl:image_retrieval_results}}
	\end{center}

\end{table*}

%% file: sections/ablation.tex
\section{Ablation Study}\label{sec:ablation}
The goal of this section is to investigate the benefits of using intra-batch hard negative mining compared to in-pair sampling used in recent CNN-based interest point detectors and descriptors~\cite{Dusmanu2019D2Net,Revaud2019R2D2}. For in-batch negative mining, we experiment with different sampling strategies: consider \textit{all} negative samples in a mini-batch, take $k$ negative samples \textit{randomly}, and consider only $top$-$k$ samples.

\customparagraph{In-pair \textit{vs.} in-batch negative mining.} To evaluate the effectiveness of the sampling procedure we conduct controlled experiments summarized in Table~\ref{tbl:ablation_batch_vs_pair}. The results are obtained by the R2D2-{$\left(\squarehvfill\right)$} model trained on the MegaDepth dataset using the AP loss function. We use the following evaluation benchmarks: sparse feature matching (HPatches), image retrieval ($\cR$Oxford5k), and visual localization (Aachen v1.1) (see Sec.~\ref{ssec:benchmarks_metrics}). As shown in Table~\ref{tbl:ablation_batch_vs_pair}, in-batch hard-negative sampling significantly outperforms in-pair mining in all metrics. We observe similar behaviour with metric learning losses, \eg, Triplet loss. The results obtained by models trained with Triplet loss are provided in supplementary.

\customparagraph{In-batch sampling: all \textit{vs.} random \textit{vs.} topK.} In contrast to \textit{pairwise} ranking losses, the AP loss is defined on a ranked list of samples $L$. Let us assume we have a mini-batch consisting of $M$ homography related image crop pairs where each crop has $p$ interest points extracted by SuperPoint~\cite{Detone18superpoint} detector. Each keypoint is associated with a feature vector and has one positive and $M\times(p-1)$ negative samples within a mini-batch. Therefore, we consider 3 ways of constructing the list $L$: take all possible negatives, randomly select $k$ samples from $M\times(p-1)$, and select $top$-$k$ negatives (the proposed strategy). For each setting, we train our unsupervised R2D2-{$\left(\squarehvfill\right)$} model on (A) and evaluate on two benchmarks: image retrieval and image-based localization. The results are presented in Table~\ref{tbl:ablation_topK_vs_all}. The proposed "top-k negative" sampling approach surpasses its counterparts with a noticeable margin. The number of hard negative samples $k$ is set to 30 after empirical evaluation for all our experiments. Interestingly, unlike Triplet loss~\cite{Mishchuk2017HardNet}, the AP loss with random sampling will \textit{not} lead to overfit, but evaluation performance is low.

\input{assets/tables/ablation_all}

%% file: assets/tables/ablation_all.tex
\begin{table}[t!]
    \begin{subtable}[t]{0.49\textwidth}
        \begin{center}
	    \resizebox{0.9\textwidth}{!}{%
		\footnotesize 		
		\begin{tabular}{c c c | c c }
		\toprule 
		\multirow{2}{*}{} & \multicolumn{2}{c|}{\multirowcell{2}{Metric}} & \multicolumn{2}{c}{Negative sampling type} \\
		& & & in-pair & in-batch \\ \midrule\midrule
		\multirow{4}{*}{\rotatebox[origin=c]{90}{$\cR$Oxford5k}} & \multirowcell{2}{mAP} & M & 55.38 & \textbf{58.69} \\
		& & H & 29.67 & \textbf{33.17} \\ \cmidrule{2-5}
		& \multirowcell{2}{mP@k $[1, 5, 10]$} & M & [94.29,	86.57, 78.43] & [\textbf{95.71}, \textbf{89.71}, \textbf{83.29}]\\
		& & H & [82.86, 54.57, 42.29] & [\textbf{85.71}, \textbf{60.29}, \textbf{45.71}]
        \\ \midrule\midrule
		\multirow{3}{*}{\rotatebox[origin=c]{90}{HPatches}} & \multirowcell{3}{MMA} & 1px & 0.239 / 0.425 / 0.332 & \textbf{0.254} / \textbf{0.439} / \textbf{0.346} \\
		& & 3px & 0.585 / 0.677 / 0.631 & \textbf{0.630} / \textbf{0.707} / \textbf{0.669} \\
		& & 5px & 0.648 / 0.742 / 0.695 & \textbf{0.706} / \textbf{0.784} / \textbf{0.745} \\[2pt] \midrule \midrule
		
		\multirow{2}{*}{\rotatebox[origin=c]{90}{Aachen}} &  & day & 87.9 / 94.2 / 97.9 &	\textbf{88.2} / \textbf{95.5} / \textbf{98.7} \\[1pt]
		& & night & 66.5 / 79.1 / 91.6 & \textbf{68.1} / \textbf{83.8} / \textbf{94.8} 
        \\[3pt] \midrule
        \end{tabular}
        }
	\end{center}
	\vspace{-15pt}
	\caption{\textbf{In-batch} \textit{vs.} \textbf{in-pair} hard negative sampling. Intra-batch negative mining performs consistently better over in-pair counterpart in all metrics. }\label{tbl:ablation_batch_vs_pair}
    \end{subtable}
        
    \begin{subtable}[h]{.49\textwidth}
        	\begin{center}
	\resizebox{0.9\textwidth}{!}{%
		\footnotesize
		\begin{tabular}{c c c | c | c c c }
		\toprule
		\multirow{2}{*}{} & \multicolumn{2}{c|}{\multirowcell{2}{Metric}} & \multirowcell{2}{Baseline} & \multicolumn{3}{c}{Negative samples in mini-batch} \\
		& & & & $all$ & $random$ & $top$-$k$ \\ \midrule\midrule
		\multirow{12}{*}{\rotatebox[origin=c]{90}{$\cR$Oxford5k}} & \multirowcell{3}{mAP} & E & $72.60$ & \underline{72.85} & $72.06$ & $\textbf{75.27}$ \\
		& & M & $54.89$ & \underline{57.06} & $56.91$ & $\textbf{57.83}$ \\
		& & H & $27.37$ & \underline{30.53} & $31.28$ & $\textbf{31.56}$ \\ \cmidrule{2-7}
		& \multirowcell{9}{mP@k $[1, 5, 10]$} & \multirowcell{3}{E} & $94.12$	& $\textbf{97.06}$ &	\underline{95.59} & $\textbf{97.06}$ \\
		& & & $80.74$ & $84.41$ & \underline{84.63} & $\textbf{88.24}$ \\
		& & & $74.12$ & $77.40$ & \underline{77.72} & $\textbf{80.44}$ \\ \cmidrule{3-7}
		& & \multirowcell{3}{M} & $91.43$ & $\textbf{95.71}$ & $94.29$ & $\textbf{95.71}$ \\
		& & & $80.57$ & \underline{88.10} & $\textbf{88.86}$ & $\textbf{88.86}$ \\
		& & & $73.86$ & \underline{81.24} & $81.00$ & $\textbf{81.41}$ \\ \cmidrule{3-7}
		& & \multirowcell{3}{H} & $61.43$ & \underline{81.43} & \underline{81.43} & $\textbf{82.86}$ \\
		& & & $47.50$ & $56.43$ & \underline{57.14} & $\textbf{59.43}$ \\
		& & & $39.79$ & $43.57$ & \underline{44.57} & $\textbf{45.43}$ \\ \midrule\midrule
		\multirowcell{3}{\rotatebox[origin=c]{90}{ Tokyo24/7}} & \multirowcell{3}{recall} & r@1 & $67.10$ & \underline{76.15} & $74.92$ & $\textbf{79.05}$ \\[2pt]
		& & r@5 & $74.20$ & \underline{81.27} & $80.63$ & $\textbf{84.13}$ \\[2pt]
		& & r@10 & $76.10$ & \underline{83.49} & $81.90$ & $\textbf{85.71}$ \\[2pt] \midrule
		\multirow{6}{*}{\rotatebox[origin=c]{90}{Aachen v1.1}} & \multirowcell{6}{$0.25m, 2^\circ$ \\ $0.5m, 5^\circ$ \\ $5m, 10^\circ$} & \multirowcell{3}{day} & \underline{87.7} & $87.5$ & $88.0$ & $\textbf{88.0}$ \\
		& & & \underline{94.7} & $94.4$ & 94.2 & $\textbf{94.8}$ \\
		& & & \textbf{98.7} & \underline{98.3} & $98.1$ & $98.2$ \\ \cmidrule{3-7}
		& & \multirowcell{3}{night} & \underline{69.6} & $\textbf{70.2}$ & $67.0$ & $\textbf{70.2}$ \\
		& & & \textbf{86.4} & \underline{85.9} & 84.3 & \textbf{86.4} \\
		& & & \underline{95.3} & \textbf{95.8} & $94.8$ & \textbf{95.8} \\ \midrule
		\multirow{6}{*}{\rotatebox[origin=c]{90}{InLoc}} & \multirowcell{6}{$0.25m, 2^\circ$ \\ $0.5m, 5^\circ$ \\ $5m, 10^\circ$} & \multirowcell{3}{DUC1} & \underline{27.8} & \textbf{28.3} & 27.3 & \underline{27.8}  \\
		& & & 42.4 & \textbf{46.0} & 43.4 & \underline{43.9} \\
		& & & \underline{54.4} & \textbf{55.6} & 53.0 & 54.0 \\ \cmidrule{3-7}
		& & \multirowcell{3}{DUC2} & 22.1 & 23.7 & \underline{26.0} & \textbf{27.5} \\
		& & & 34.4 & 35.1 & \underline{36.6} & \textbf{38.2} \\
		& & & 42.7 & 40.5 & \underline{44.3} & \textbf{49.6} \\ \midrule
		\end{tabular}
		
	}
	
	\end{center}
	\vspace{-15pt}
	\caption{\textbf{Intra-batch negative sampling}. For a listwise loss, the $top$-$k$ sampling strategy leads to a better performance over its counterparts in almost all metrics. The best score is in bold and the second best score is underlined.}\label{tbl:ablation_topK_vs_all}
    \end{subtable}
    \vspace{-5pt}
    \caption{\textbf{Ablation study.} We analyze the influence of different design choices of the proposed approach. See Sec.~\ref{sec:ablation} for more details.}
    \vspace{-15pt}
\end{table}

%% file: sections/conclusion.tex
\section{Conclusion}
In this work, we have presented a self-supervised local descriptor learning framework. We introduced three main contributions: (1) a coarse-to-fine method to efficiently sample hard-negative local descriptors by leveraging visual image descriptors, (2) the modified version of AP loss which can carefully handle imbalance between the number of positive and negative keypoint descriptors and (3) photorealistic image stylization that combined with synthetic color augmentation can significantly enhance the robustness of learned local descriptors to illumination changes. The proposed improvements together result in a self-supervised approach demonstrates favourable performance on a wide range of geometric tasks compared to strong fully- and weakly-supervised baseline models.

%% file: supp/supp_main.tex
\section{Overview}
In this supplementary material, we provide additional results and visualization of the proposed self-supervised framework for local image descriptos learning. In Sec.~\ref{sec:training_details}, we provide the details of the training data and training procedure. Sec.~\ref{sec:rdnim_results} shows the evaluation on the recently proposed Rotated Day-Night Image Matching dataset (RDNIM)~\cite{Pautrat2020LISRD} representing strong illumination and viewpoint changes. To verify generalization performance of our method, we present ablation study on different detectors in Sec.~\ref{sec:ablation_detectors_triplet}. In Sec.~\ref{sec:qualitative_results}, we demonstrate more qualitative results of matches produced by our method on outdoor and indoor benchmarks. 

\section{Training details}\label{sec:training_details}
In this work, we use three large-scale outdoor datasets: MegaDepth (M)~\cite{Li2018Megadepth}, Phototourism (P), and database images of Aachen v1.1 (A). For M and P, the official training and validation splits have been utilized to optimize the proposed approach presented in the main part in Figure~1. All models are trained in an end-to-end fashion with the Adam optimizer~\cite{Kingma2015AdamSolver} and an initial learning rate of $10^{-3}$ and $10^{-4}$ for unsupervised versions of R2D2 and CAPS models respectively. We use the {$\left(\squarehvfill\right)$} notation for the models trained on synthetic homographies with color augmentations while the {$\left(\squarehvfill,\squarerightblack\right)$} encodes the models trained on images with color augmentations and stylization. The training was performed on a single GeForce RTX 2080Ti GPU. The style categories and image stylization utilized during training are illustrated in Figure~\ref{fig:supp_style_images}.

\customparagraph{Synthetic color augmentation.} To improve generalization performance of the proposed approach, we apply synthetic color augmentation separately on the source and target image crops. Specifically, we utilize histogram equalization (CLAHE) and add per-pixel Gaussian noise and Gaussian blur along with color augmentation in contrast, brightness, hue, and saturation. The full set of color augmentations is presented in Figure~\ref{fig:supp_augmentations}.

\input{supp/assets/tables/res_rdnim}

\section{Evaluation on RDNIM}\label{sec:rdnim_results}
In order to verify the robustness of local image descriptors to severe appearance and viewpoint changes, Pautrat~\etal~\cite{Pautrat2020LISRD} propose the RDNIM dataset. The dataset originates from the DNIM dataset -- a subset of the AMOS database~\cite{Jacobs2007AMOS} represents a large number of images taken at regular time intervals by outdoor webcams with fixed positions and orientations. The DNIM dataset consists of sequences, with a total of 1722 day-time and night-time images. Pautrat~\etal~\cite{Pautrat2020LISRD} provide two benchmarks, where the images of each sequence are paired with either the day-time or the night-time image. The images are then augmented with homographic warps including rotations similarly to~\cite{Detone18superpoint}. 

The evaluation of the state-of-the-art local descriptors with SuperPoint detector is summarized in Table~\ref{tbl:supp_rdnim_results}. For LISRD, in addition to the results presented in~\cite{Pautrat2020LISRD}, we also evaluate one of the models publicly available at \url{https://github.com/rpautrat/LISRD}. We refer to this model as LISRD*. The proposed models perform better than the fully- and weakly-supervised counterparts (R2D2~\cite{Revaud2019R2D2} and CAPS~\cite{Wang2020CAPS}, respectively) in all metrics. The models have been trained on stylized images with synthetic color augmentation,~\ie, $\left(\squarehvfill,\squarerightblack\right)$. Notably, using larger model (CAPS) and visual image descriptors (-gd) leads to better performance which is consistent with the results obtained for image retrieval and camera relocalization benchmarks presented in the main part.

\input{supp/assets/tables/res_aachen_triplet}

\input{supp/assets/tables/res_aachen}

\section{Ablation study}\label{sec:ablation_detectors_triplet}
\customparagraph{Triplet Loss.} In addition to the results presented in our main work, we verify the proposed ideas of hard-negative sampling and photo-realistic image stylization using Triplet loss. For in-pair sampling, the closest non-matching descriptor is selected for each positive-anchor pair from a set of descriptors extracted from an input image pair. To create our supervised baseline, we adapt the model by Revaud~\etal~\cite{Revaud2019R2D2} and train it with Triplet loss using ground-truth pixel correspondences produced by optical flow. We evaluate the baselines and our models on the Aachen Day-Night v1.1~\cite{Zhang2020AachenV11} dataset and report the percentage of correctly localized queries under specific error thresholds. The results are presented in Table~\ref{tbl:supp_aachen_triplet_results}. Interestingly, the triplet margin loss and the list-wise loss discussed in the main part behave very similarly. The in-batch hard negative sampling leads to better results over in-pair mining. Utilizing both synthetic color augmentation and image stylization is very powerful and can significantly improve localization performance.

\customparagraph{Detectors.} In order to analyze generalization performance of the proposed method, we evaluate our local descriptors on different keypoint detectors. Specifically, we train our models with SuperPoint keypoints but utilize SIFT~\cite{Lowe2004SIFT} and D2-Net~\cite{Dusmanu2019D2Net} detectors during evaluation. We consider the problem of visual localization as our benchmark and report the number of correctly localized query images of Aachen Day-Night~\cite{Zhang2020AachenV11} under different settings in Table~\ref{tbl:supp_aachen_loc}. As shown in Table~\ref{tbl:supp_aachen_loc}, our models are not limited to SuperPoint detector but perform favorably with other detectors. The models leveraging visual image descriptors for hard-negative mining during training demonstrate better generalization performance and localization results (Table~\ref{tbl:supp_loc_aachen_vis_descs}, D2-Net detector).

\input{supp/assets/images/fig_augmentations}

\section{Qualitative results}\label{sec:qualitative_results}
We visualize the correspondences estimated by our pipeline (with CAPS backbone network) and its weakly-supervised strong counterpart on the image pairs of Aachen Day-Night~\cite{Zhang2020AachenV11} in Figure~\ref{fig:supp_aachen_caps} and of InLoc~\cite{Taira2018InLoc} in Figure~\ref{fig:supp_inloc_caps}. The matches are verified by using the \texttt{findFundamentalMatrix} function with a RANSAC threshold of 1~\cite{Zhou2021Patch2Pix}. For a randomly selected query image, we choose the database image with the most inlier correspondences verified by the camera pose solver. Figure~\ref{fig:supp_inloc_r2d2} shows qualitative results on InLoc queries obtained by the model based on R2D2 architecture. The correspondences are illustrated in green.

\input{supp/assets/images/fig_styles_examples}
\input{supp/assets/images/fig_aachen_caps}
\input{supp/assets/images/fig_inloc_caps}
\input{supp/assets/images/fig_inloc_r2d2}

%% file: supp/assets/tables/res_rdnim.tex
\begin{table}[t!]
	\begin{center}
	\resizebox{.45\textwidth}{!}{%
		\footnotesize
		\begin{tabular}{c l | c c c | c c c }
		\toprule
		\multirow{2}{*}{} & \multirowcell{2}{Method} & \multicolumn{3}{c|}{Day reference} & \multicolumn{3}{c}{Night reference} \\
		& & \textEta & Precision & Recall & \textEta & Precision & Recall \\
		\midrule\midrule
		& LISRD~\cite{Pautrat2020LISRD} & 0.198 & 0.291 & 0.317 & 0.262 & 0.371 & 0.384\\
		& LISRD* & \textbf{0.358} & \textbf{0.433} & \textbf{0.526} & \textbf{0.442} & \textbf{0.541} & \textbf{0.621}\\
		& HardNet~\cite{Mishchuk2017HardNet} & 0.249 & 0.225 & 0.224 & 0.325 & 0.359 & 0.365\\
		& SOSNet~\cite{Tian2019SOSNET} & 0.226 & 0.218 & 0.226 & 0.252 & 0.288 & 0.296\\
		& SuperPoint~\cite{Detone18superpoint} & 0.178 & 0.191 & 0.214 & 0.235 & 0.259 & 0.296\\
		\midrule
		& \textcolor{Magenta}{R2D2}~\cite{Revaud2019R2D2} & 0.215 & 0.219 & 0.216 & 0.254 & 0.289 & 0.280\\
		& \textcolor{Cyan}{CAPS}~\cite{Wang2020CAPS} & 0.159 & 0.258 & 0.278 & 0.212 & 0.308 & 0.310\\
		\midrule\midrule
		\multirow{8}{*}{\rotatebox[origin=c]{90}{Unsupervised (ours)}} &
		R2D2-[M] & 0.233 & 0.238 & 0.244 & \textcolor{Magenta}{\underline{0.281}} & 0.312 & 0.314 \\
		& R2D2-[M+P] & 0.222 & \textcolor{Magenta}{\underline{0.245}} & \textcolor{Magenta}{\underline{0.253}} & 0.274 & \textcolor{Magenta}{\underline{0.317}} & \textcolor{Magenta}{\underline{0.322}} \\
		& R2D2-[M]-selfgd & \textcolor{Magenta}{\underline{0.237}} & 0.229 & 0.228 & 0.278 & 0.304 & 0.303 \\
		& R2D2-[M]-gd & 0.229 & 0.230 & 0.239 & 0.274 & 0.307 & 0.311 \\ \cmidrule{2-8}
		& CAPS-[M] & 0.180 & \textcolor{Cyan}{\underline{0.267}} & 0.256 & 0.250 & \textcolor{Cyan}{\underline{0.337}} & 0.316 \\
		& CAPS-[M+P] & 0.193 & \textcolor{Cyan}{\underline{0.267}} & 0.254 & 0.250 & 0.336 & 0.314 \\
		& CAPS-[M]-selfgd & \textcolor{Cyan}{\underline{0.215}} & 0.263 & \textcolor{Cyan}{\underline{0.282}} & \textcolor{Cyan}{\underline{0.271}} & 0.331 & 0.348 \\
		& CAPS-[M]-gd & 0.209 & \textcolor{Cyan}{\underline{0.267}} & \textcolor{Cyan}{\underline{0.282}} & 0.262 & 0.336 & \textcolor{Cyan}{\underline{0.350}} \\
		\end{tabular}
		
	}
	\end{center}
	\caption{\textbf{Evaluation results on RDNIM}~\cite{Pautrat2020LISRD} We follow~\cite{Pautrat2020LISRD} and report the accuracy of homography estimation, precision and recall for error thresholds of 3 pixels. The best score between supervised (weakly-) baselines,~\ie, \textcolor{Magenta}{R2D2} and \textcolor{Cyan}{CAPS}, and the proposed unsupervised methods for each category is \underline{underlined}. The overall best score is in \textbf{bold}.}\label{tbl:supp_rdnim_results}
\end{table}

%% file: supp/assets/tables/res_aachen_triplet.tex
\begin{table}[t!]
    \begin{center}
    \resizebox{.49\textwidth}{!}{%
	\footnotesize 		
	\begin{tabular}{l | c c | c c c | c c c}
	\toprule
	\multirowcell{4}{Method} & \multirowcell{4}{Supervision} & \multirowcell{4}{Hard-negative\\sampling} & \multicolumn{6}{c}{Aachen v1.1}\\
	& & & \multicolumn{6}{c}{\% localized queries} \\
	& & & \multicolumn{3}{c|}{Day (824 images)} & \multicolumn{3}{c}{Night (191 images)} \\
	& & & $0.25m, 2^\circ$ & $0.5m, 5^\circ$ & $5m, 10^\circ$ & $0.25m, 2^\circ$ & $0.5m, 5^\circ$ & $5m, 10^\circ$ \\ \midrule\midrule
	R2D2~\cite{Revaud2019R2D2}-[A+R] & OF & in-pair & 86.5 & 95.0 & 98.2 & 69.1 & 84.5 & 93.2 \\
	R2D2~\cite{Revaud2019R2D2}-[A+R] & OF & in-batch & 88.6 & 95.5 & 98.7 & 71.2 & 86.4 & 94.8 \\ \midrule
	R2D2-{$\left(\squarehvfill\right)$}-[A] & - & in-batch & 87.9 & 94.5 & 98.4 & 63.9 & 79.1 & 90.6 \\
	R2D2-{$\left(\squarehvfill,\squarerightblack\right)$}-[A] & - & in-batch & 87.1 & 94.3 & 98.4 & 68.1 & 84.3 & 92.7 \\
	R2D2-{$\left(\squarehvfill,\squarerightblack\right)$}-[M+P] & - & in-batch & 88.1 & 95.1 & 98.6 & 72.1 & 88.2 & 95.4 \\
	\end{tabular}
	}
	\end{center}
	\caption{\textbf{Localization with Triplet loss.} The percentage of correctly localized queries of Aachen Day-Night v1.1~\cite{Zhang2020AachenV11} under specific thresholds. \textbf{Legend}: \textit{Supervision}: OF~--~ground-truth pixel correspondences based on optical flow~\cite{Revaud2019R2D2}. The training datasets are given in $\left[.\right]$ parentheses: A~--~Aachen database images; R~--~Random images from the Internet~\cite{Revaud2019R2D2}; M~--~MegaDepth dataset; P~--~Phototourism dataset.}\label{tbl:supp_aachen_triplet_results}
	
\end{table}

%% file: supp/assets/tables/res_aachen.tex
\begin{table*}[t!]
    \begin{subtable}[h]{\textwidth} 
        \begin{center}
	    \resizebox{\textwidth}{!}{%
		\footnotesize
		\begin{tabular}{c c | c c c | c c c | c c c}
		\toprule
		\multirow{2}{*}{} & \multirowcell{2}{Error\\threshold} & \multicolumn{3}{|c}{SIFT detector} & \multicolumn{3}{|c}{SuperPoint detector} & \multicolumn{3}{|c}{D2-Net detector} \\
		& & SIFT & Ours (R2D2) & Ours (CAPS) & SuperPoint &  Ours (R2D2) & Ours (CAPS) & D2-Net (SS\footnotemark) & Ours (R2D2) & Ours (CAPS) \\
		\hline\hline
		\multirow{3}{*}{\rotatebox[origin=c]{90}{day}} & $0.25m, 2^\circ$ & 76.2 & \textbf{80.0} & 77.3 & 86.8 & \textbf{87.4} & 85.1 & \textbf{84.0} & 83.2 & 79.2 \\
		& $0.5m, 5^\circ$ & 84.3 & \textbf{88.6} & 88.5 & 94.1 & \textbf{94.7} & 93.2 & 92.1 & \textbf{92.2} & 89.0 \\
		& $5m, 10^\circ$ & 91.4 & 94.9 & \textbf{95.4} & \textbf{98.3} & \textbf{98.3} & 97.8 & \textbf{97.8} & 97.3 & 97.3 \\ \hline
		\multirow{3}{*}{\rotatebox[origin=c]{90}{night}} & $0.25m, 2^\circ$ & 52.1 & 56.5 & \textbf{61.4} & 68.1 & \textbf{72.3} & 71.7 & \textbf{73.8} & 68.1 & 67.0 \\
		& $0.5m, 5^\circ$ & 63.3 & 71.2 & \textbf{77.1} & 85.9 & \textbf{88.5} & 87.4 & \textbf{92.1} & 87.3 & 85.3 \\
		& $5m, 10^\circ$ & 74.5 & 84.8 & \textbf{93.3} & 96.9 & 97.4 & \textbf{97.9} & \textbf{98.4} & 97.2 & 97.9 \\
		
		\end{tabular}
        }
        
	\end{center}
	\caption{Localization performance of the proposed keypoint descriptor with different detectors}\label{tbl:supp_loc_aachen_pure}
    \end{subtable}
    \begin{subtable}[h]{\textwidth} 
        \begin{center}
	    \resizebox{\textwidth}{!}{%
		\footnotesize 		
		\begin{tabular}{c c | c c c c c | c c c c c | c c c c c}
		\toprule
		\multirow{3}{*}{} & \multirowcell{3}{Error\\threshold} & \multicolumn{5}{|c}{SIFT detector} & \multicolumn{5}{|c}{SuperPoint detector} & \multicolumn{5}{|c}{D2-Net detector} \\
		& & \multirowcell{2}{SIFT} & \multicolumn{2}{c}{Ours (R2D2)} & \multicolumn{2}{c|}{Ours (CAPS)} & \multirowcell{2}{SuperPoint} &  \multicolumn{2}{c}{Ours (R2D2)} & \multicolumn{2}{c|}{Ours (CAPS)} & \multirowcell{2}{D2-Net (SS)} & \multicolumn{2}{c}{Ours (R2D2)} & \multicolumn{2}{c}{Ours (CAPS)} \\
		& & & selfgd & gd & selfgd & gd & & selfgd & gd & selfgd & gd & & selfgd & gd & selfgd & gd \\
		\hline\hline
		\multirow{3}{*}{\rotatebox[origin=c]{90}{day}} & $0.25m, 2^\circ$ & 76.2 & 79.2 & \textbf{79.9} & 79.1 & 78.5 & 86.8 & \textbf{88.1} & 87.5 & 86.9 & 87.0 & 84.0 & 83.4 & \textbf{85.0} & 82.6 & 82.9 \\
		& $0.5m, 5^\circ$ & 84.3 & 88.1 & 88.2 & \textbf{89.1} & 88.8 & 94.1 & 94.8 & \textbf{94.9} & 93.8 & 93.8 & 92.1 & 92.5 & \textbf{92.6} & 90.5 & 91.1 \\
		& $5m, 10^\circ$ & 91.4 & 93.6 & 94.4 & 95.4 & \textbf{95.6} & \textbf{98.3} & 98.1 & \textbf{98.3} & 98.1 & \textbf{98.3} & 97.8 & 97.5 & 97.8 & 97.5 & \textbf{98.1} \\ \hline
		\multirow{3}{*}{\rotatebox[origin=c]{90}{night}} & $0.25m, 2^\circ$ & 52.1 & 49.7 & 52.4 & 64.9 & \textbf{66.0} & 68.1 & 71.2 & 71.7 & 71.7 & \textbf{73.8} & \textbf{73.8} & 69.6 & 70.7 & 71.2 & 69.1 \\
		& $0.5m, 5^\circ$ & 63.3 & 67.5 & 71.2 & \textbf{83.8} & 81.7 & 85.9 & 88.0 & 86.4 & \textbf{89.0} & \textbf{89.0} & \textbf{92.1} & 87.4 & 88.0 & 84.8 & 87.4\\
		& $5m, 10^\circ$ & 74.5 & 78.5 & 84.8 & \textbf{95.8} & 94.8 & 96.9 & 95.8 & 96.9 & \textbf{97.4} & \textbf{97.4} & \textbf{98.4} & 97.4 & 97.4 & 97.9 & \textbf{98.4} \\
		
		\end{tabular}
        }
	\end{center}
	\caption{Localization performance of the proposed method utilizing \textit{visual image descriptors} during training.}\label{tbl:supp_loc_aachen_vis_descs}
    \end{subtable}
    \caption{\textbf{Visual localization performance on Aachen Day-Night v1.1}~\cite{Zhang2020AachenV11}. To verify generalization performance, we evaluate the proposed learned local descriptor on Aachen v1.1 using different keypoint detectors. The overall best score for each detector is in \textbf{bold}. }\label{tbl:supp_aachen_loc}
    
\end{table*}
\footnotetext[1]{We evaluate the single scale model only due to technical restrictions of our hardware}

%% file: supp/assets/images/fig_augmentations.tex
\begin{figure}[t!]
  \centering
    \includegraphics[width=1\linewidth]{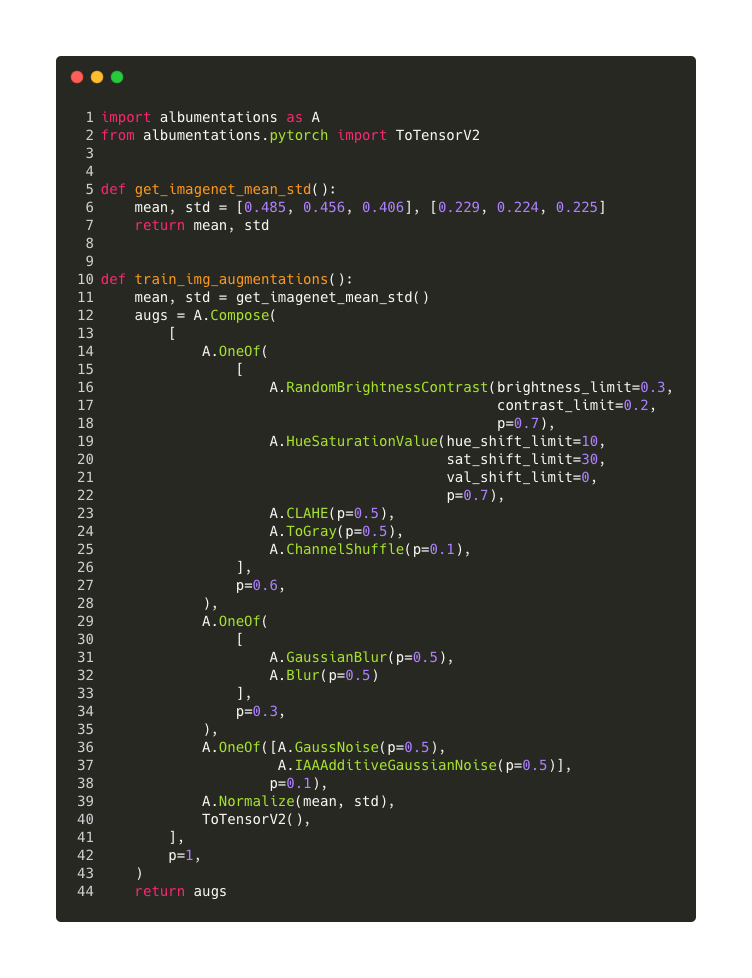}
 \caption{\textbf{Color augmentation} utilized in the proposed pipeline. We used Albumentations~\cite{buslaev2020Albumentations} library to generate random color transforms of training images}  
 \label{fig:supp_augmentations}
\end{figure}

%% file: supp/assets/images/fig_styles_examples.tex
\begin{figure*}[t!]
  \centering
    \includegraphics[width=1\linewidth]{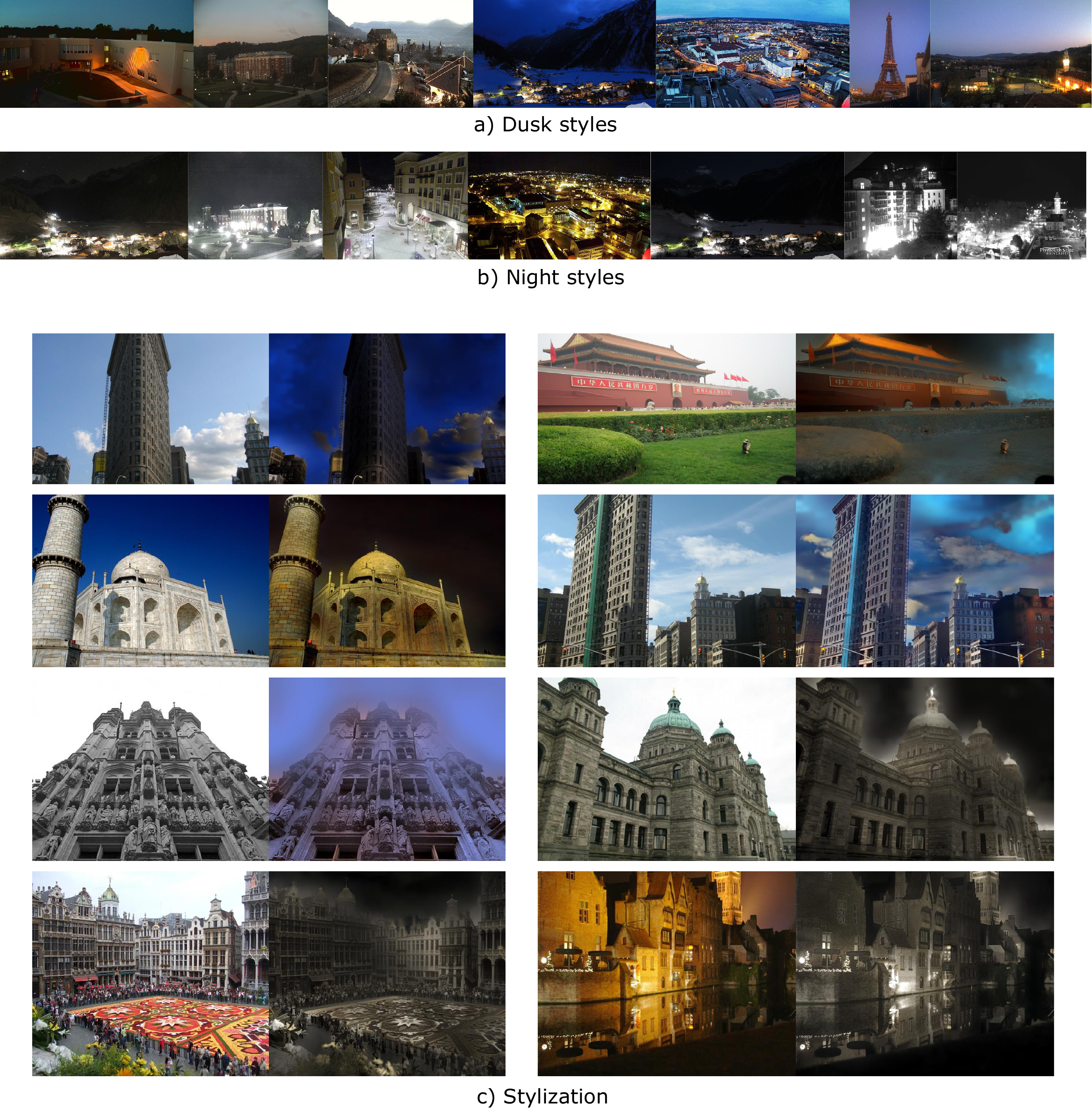}
 \caption{\textbf{The set of images used for photo-realistic stylization}. In contrast to~\cite{Revaud2019R2D2}, we use two style categories, \ie Dusk and Night, and 7 style examples for each category. We randomly sample a style image from these two categories and apply it to each particular image from the training dataset.}  
 \label{fig:supp_style_images}
\end{figure*}

%% file: supp/assets/images/fig_aachen_caps.tex
\begin{figure*}[t!]
  \centering
    \includegraphics[width=1\linewidth]{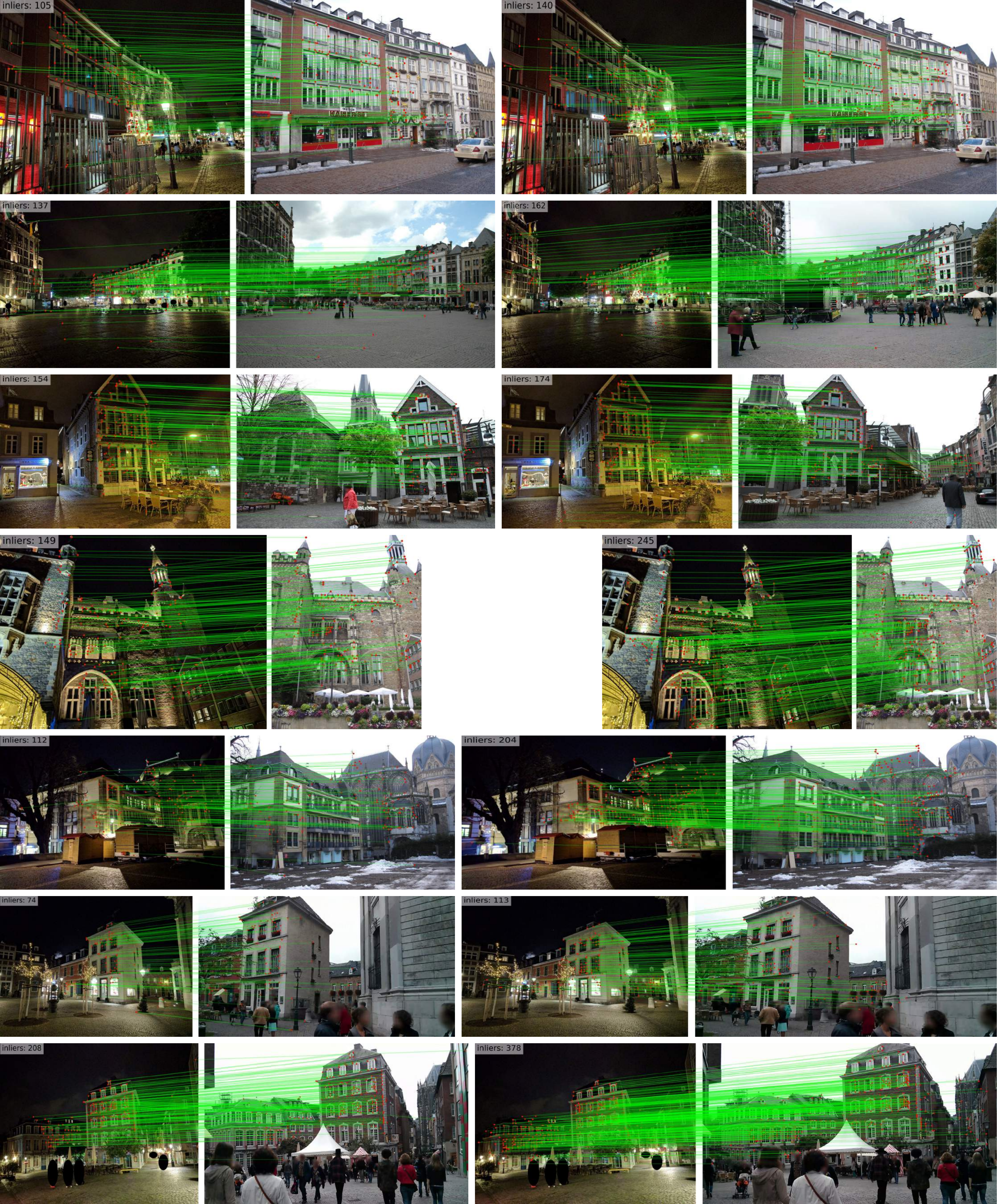}
 \caption{\textbf{Example correspondences} produced by our method (left) and its weakly-supervised counterpart,~\ie CAPS~\cite{Wang2020CAPS} (right), on night-time queries of Aachen Day-Night (v1.1)~\cite{Zhang2020AachenV11}}  
 \label{fig:supp_aachen_caps}
\end{figure*}

%% file: supp/assets/images/fig_inloc_caps.tex
\begin{figure*}[t!]
  \centering
    \includegraphics[width=1\linewidth]{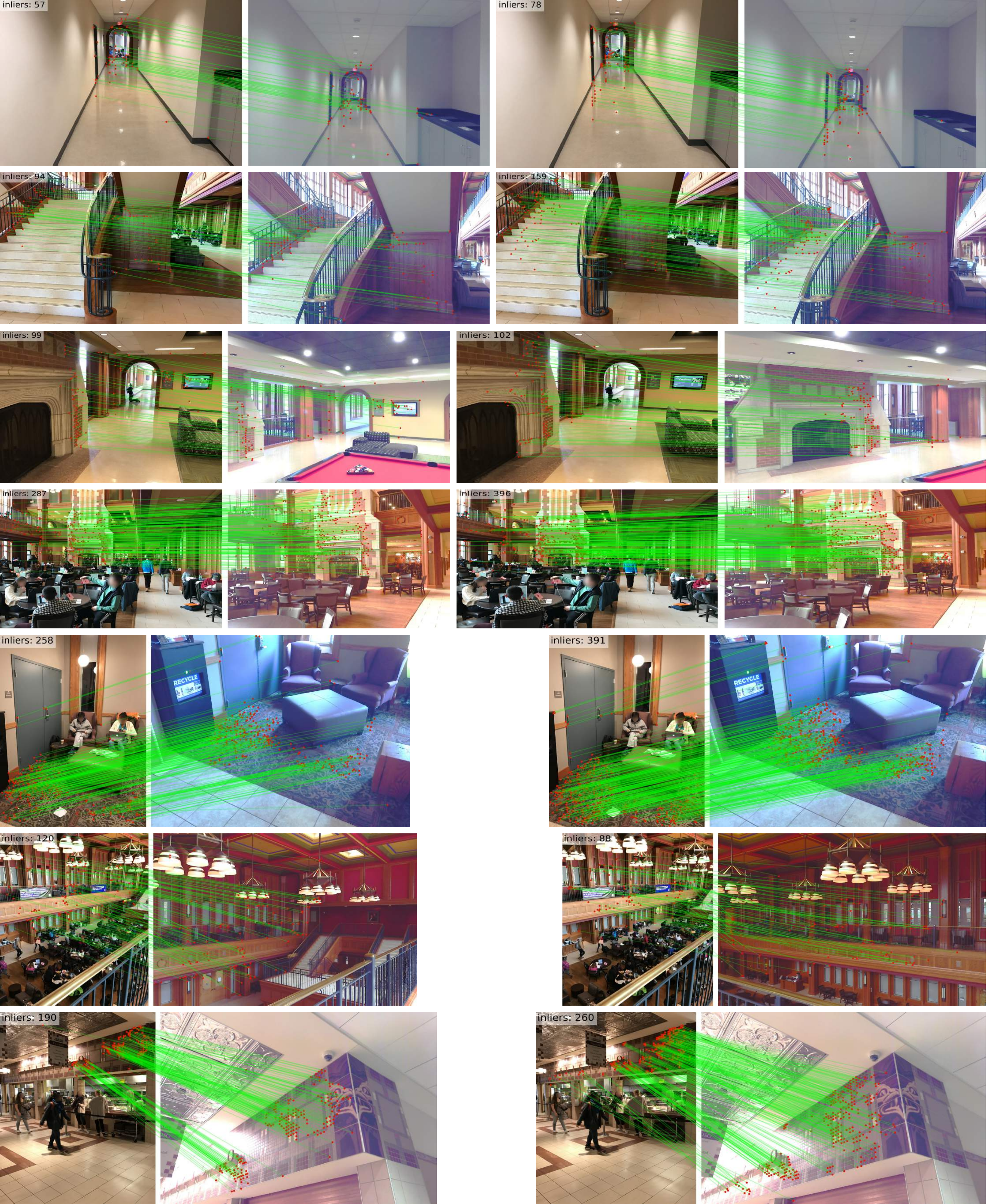}
 \caption{\textbf{Example correspondences} produced by our method (left) and its weakly-supervised counterpart,~\ie CAPS~\cite{Wang2020CAPS} (right), on queries of InLoc~\cite{Taira2018InLoc}}  
 \label{fig:supp_inloc_caps}
\end{figure*}

%% file: supp/assets/images/fig_inloc_r2d2.tex
\begin{figure*}[t!]
  \centering
    \includegraphics[width=1\linewidth]{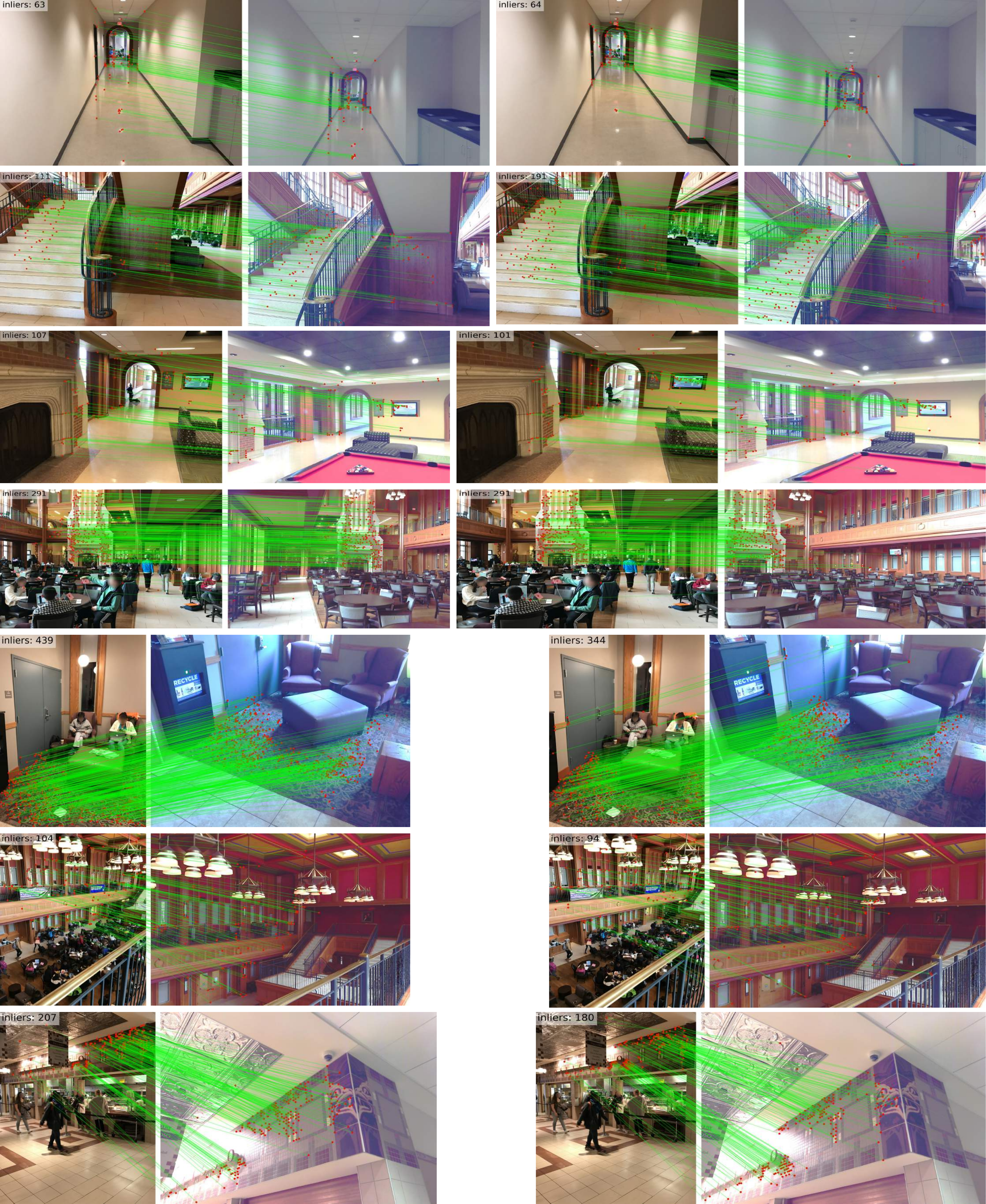}
 \caption{\textbf{Example correspondences} produced by our method (left) and its supervised counterpart,~\ie R2D2~\cite{Revaud2019R2D2} (right), on queries of InLoc~\cite{Taira2018InLoc}}  
 \label{fig:supp_inloc_r2d2}
\end{figure*}

%% file: main.bbl
\begin{thebibliography}{10}\itemsep=-1pt

\bibitem{Anoosheh2020S2Night2Day}
Asha Anoosheh, Torsten Sattler, Radu Timofte, Marc Pollefeys, and Luc~Van Gool.
\newblock Night-to-day image translation for retrieval-based localization.
\newblock In {\em International Conference on Robotics and Automation (ICRA)},
  pages 5958--5964, 2019.

\bibitem{Arandjelovic2016NetVLAD}
Relja Arandjelovi{\'c}, Petr Gronat, Akihiko Torii, Tomas Pajdla, and Josef
  Sivic.
\newblock {NetVLAD}: {CNN} architecture for weakly supervised place
  recognition.
\newblock In {\em Proceedings of the IEEE/CVF Conference on Computer Vision and
  Pattern Recognition (CVPR)}, pages 5297--5307, 2016.

\bibitem{bachman_2019}
Philip Bachman, R~Devon Hjelm, and William Buchwalter.
\newblock Learning representations by maximizing mutual information across
  views.
\newblock In {\em Advances in Neural Information Processing Systems (NeurIPS)},
  volume~32, pages 15535--15545. Curran Associates, Inc., 2019.

\bibitem{Baik2020DomainAdaptation}
Sungyong Baik, Hyo~Jin Kim, Tianwei Shen, Eddy Ilg, Kyoung~Mu Lee, and Chris
  Sweeney.
\newblock Domain adaptation of learned features for visual localization.
\newblock In {\em Proceedings of the British Machine Vision Conference (BMVC)},
  2020.

\bibitem{Balntas2017HPatches}
Vassileios Balntas, Karel Lenc, Andrea Vedaldi, and Krystian Mikolajczyk.
\newblock {HPatches}: A benchmark and evaluation of handcrafted and learned
  local descriptors.
\newblock In {\em Proceedings of the IEEE/CVF Conference on Computer Vision and
  Pattern Recognition (CVPR)}, pages 3852--3861, 2017.

\bibitem{Balntas2016Localdesc}
Vassileios Balntas, Edgar Riba, Daniel Ponsa, and Krystian Mikolajczyk.
\newblock Learning local feature descriptors with triplets and shallow
  convolutional neural networks.
\newblock In {\em Proceedings of the British Machine Vision Conference (BMVC)},
  2016.

\bibitem{Barroso2019KeyNet}
Axel Barroso-Laguna, Edgar Riba, Daniel Ponsa, and Krystian Mikolajczyk.
\newblock {Key.Net}: Keypoint detection by handcrafted and learned {CNN}
  filters.
\newblock In {\em Proceedings of the IEEE/CVF International Conference on
  Computer Vision (ICCV)}, pages 5835--5843, 2019.

\bibitem{Bay2006SURF}
Herbert Bay, Tinne Tuytelaars, and Luc~Van Gool.
\newblock {SURF}: Speeded up robust features.
\newblock In {\em Proceedings of the European Conference on Computer Vision
  (ECCV)}, pages 404--417. Springer Berlin Heidelberg, 2006.

\bibitem{Bhowmik2020ReinforcedPoints}
Aritra Bhowmik, Stefan Gumfold, Carsten Rother, and Eric Brachmann.
\newblock Reinforced feature points: Optimizing feature detection and
  description for a high-level task.
\newblock In {\em Proceedings of the IEEE/CVF Conference on Computer Vision and
  Pattern Recognition (CVPR)}, pages 4947--4956, 2020.

\bibitem{buslaev2020Albumentations}
Alexander Buslaev, Vladimir~I. Iglovikov, Eugene Khvedchenya, Alex Parinov,
  Mikhail Druzhinin, and Alexandr~A. Kalinin.
\newblock Albumentations: Fast and flexible image augmentations.
\newblock {\em Information}, 11(2), 2020.

\bibitem{Calonder2010BRIEF}
Michael Calonder, Vincent Lepetit, Christoph Strecha, and Pascal Fua.
\newblock {BRIEF}: Binary robust independent elementary features.
\newblock In {\em Proceedings of the European Conference on Computer Vision
  (ECCV)}, pages 778--792. Springer Berlin Heidelberg, 2010.

\bibitem{chen2020SimCLR}
Ting Chen, Simon Kornblith, Mohammad Norouzi, and Geoffrey Hinton.
\newblock A simple framework for contrastive learning of visual
  representations.
\newblock In {\em Proceedings of the International Conference on Machine
  Learning (ICML)}, 2020.

\bibitem{Christiansen2019Unsuperpoint}
Peter~Hviid Christiansen, Mikkel~Fly Kragh, Yury Brodskiy, and Henrik Karstoft.
\newblock {UnsuperPoint}: End-to-end unsupervised interest point detector and
  descriptor.
\newblock {\em ArXiv preprint arXiv:1907.04011}, 2019.

\bibitem{debiased}
Ching-Yao Chuang, Joshua Robinson, Yen-Chen Lin, Antonio Torralba, and Stefanie
  Jegelka.
\newblock Debiased contrastive learning.
\newblock In {\em Advances in Neural Information Processing Systems (NeurIPS)},
  volume~33, pages 8765--8775. Curran Associates, Inc., 2020.

\bibitem{Cubuk2019RandAugment}
Ekin~D. Cubuk, Barret Zoph, Jonathon Shlens, and Quoc~V. Le.
\newblock {RandAugment}: Practical automated data augmentation with a reduced
  search space.
\newblock In {\em Proceedings of the IEEE/CVF Conference on Computer Vision and
  Pattern Recognition (CVPR) Workshops}, pages 3008--3017, 2019.

\bibitem{Deng2009ImageNet}
Jia Deng, Wei Dong, Richard Socher, Li-Jia Li, Kai Li, and Li Fei-Fei.
\newblock {ImageNet}: A large-scale hierarchical image database.
\newblock In {\em Proceedings of the IEEE/CVF Conference on Computer Vision and
  Pattern Recognition (CVPR)}, pages 248--255, 2009.

\bibitem{Detone18superpoint}
Daniel DeTone, Tomasz Malisiewicz, and Andrew Rabinovich.
\newblock {SuperPoint}: Self-supervised interest point detection and
  description.
\newblock In {\em Proceedings of the IEEE/CVF Conference on Computer Vision and
  Pattern Recognition (CVPR) Workshops}, pages 337--349, 2018.

\bibitem{doersch_2015}
Carl Doersch, Abhinav Gupta, and Alexei~A. Efros.
\newblock Unsupervised visual representation learning by context prediction.
\newblock In {\em Proceedings of the IEEE/CVF International Conference on
  Computer Vision (ICCV)}, pages 1422--1430, 2015.

\bibitem{dosovitskiy_2014}
Alexey Dosovitskiy, Jost~Tobias Springenberg, Martin Riedmiller, and Thomas
  Brox.
\newblock Discriminative unsupervised feature learning with convolutional
  neural networks.
\newblock In {\em Advances in Neural Information Processing Systems (NIPS)},
  volume~27, pages 766--774. Curran Associates, Inc., 2014.

\bibitem{Dusmanu2019D2Net}
Mihai Dusmanu, Ignacio Rocco, Tomas Pajdla, Marc Pollefeys, Josef Sivic,
  Akihiko Torii, and Torsten Sattler.
\newblock {D2-Net}: A trainable cnn for joint detection and description of
  local features.
\newblock In {\em Proceedings of the IEEE/CVF Conference on Computer Vision and
  Pattern Recognition (CVPR)}, pages 8084--8093, 2019.

\bibitem{Fang2019InstaBoost}
Hao-Shu Fang, Jianhua Sun, Runzhong Wang, Minghao Gou, Yong-Lu Li, and Cewu Lu.
\newblock {InstaBoost}: Boosting instance segmentation via probability map
  guided copy-pasting.
\newblock In {\em Proceedings of the IEEE/CVF International Conference on
  Computer Vision (ICCV)}, pages 682--691, 2019.

\bibitem{Germain2020S2DNet}
Hugo Germain, Guillaume Bourmaud, and Vincent Lepetit.
\newblock {S2DNet}: Learning image features for accurate sparse-to-dense
  matching.
\newblock In {\em Proceedings of the European Conference on Computer Vision
  (ECCV)}, pages 626--643. Springer International Publishing, 2020.

\bibitem{Goodfellow2014GANs}
Ian Goodfellow, Jean Pouget-Abadie, Mehdi Mirza, Bing Xu, David Warde-Farley,
  Sherjil Ozair, Aaron Courville, and Yoshua Bengio.
\newblock Generative adversarial nets.
\newblock In {\em Advances in Neural Information Processing Systems (NIPS)},
  volume~27, pages 2672--2680. Curran Associates, Inc., 2014.

\bibitem{Gordo2017Retrieval}
Albert Gordo, Jon Almaz{\'a}n, Jerome Revaud, and Diane Larlus.
\newblock End-to-end learning of deep visual representations for image
  retrieval.
\newblock {\em International Journal on Computer Vision (IJCV)},
  124(2):237--254, 2017.

\bibitem{he2020moco}
Kaiming He, Haoqi Fan, Yuxin Wu, Saining Xie, and Ross Girshick.
\newblock Momentum contrast for unsupervised visual representation learning.
\newblock In {\em Proceedings of the IEEE/CVF Conference on Computer Vision and
  Pattern Recognition (CVPR)}, pages 9726--9735, 2020.

\bibitem{He2018LocDesc}
Kun He, Yan Lu, and Stan Sclaroff.
\newblock Local descriptors optimized for average precision.
\newblock In {\em Proceedings of the IEEE/CVF Conference on Computer Vision and
  Pattern Recognition (CVPR)}, pages 596--605, 2018.

\bibitem{He2016ResNet}
Kaiming He, Xiangyu Zhang, Shaoqing Ren, and Jian Sun.
\newblock Deep residual learning for image recognition.
\newblock In {\em Proceedings of the IEEE/CVF Conference on Computer Vision and
  Pattern Recognition (CVPR)}, pages 770--778, 2016.

\bibitem{heinly2015_reconstructing_the_world}
Jared Heinly, Johannes~Lutz Sch\"{o}nberger, Enrique Dunn, and Jan-Michael
  Frahm.
\newblock Reconstructing the world* in six days.
\newblock In {\em Proceedings of the IEEE/CVF Conference on Computer Vision and
  Pattern Recognition (CVPR)}, pages 3287--3295, 2015.

\bibitem{Jacobs2007AMOS}
Nathan Jacobs, Nathaniel Roman, and Robert Pless.
\newblock Consistent temporal variations in many outdoor scenes.
\newblock In {\em Proceedings of the IEEE/CVF Conference on Computer Vision and
  Pattern Recognition (CVPR)}, pages 1--6, 2007.

\bibitem{Jiang2021COTR}
Wei Jiang, Eduard Trulls, Jan Hosang, Andrea Tagliasacchi, and Kwang~Moo Yi.
\newblock {COTR}: Correspondence transformer for matching across images.
\newblock In {\em Proceedings of the IEEE/CVF International Conference on
  Computer Vision (ICCV)}, pages 6207--6217, 2021.

\bibitem{kalantidis}
Yannis Kalantidis, Mert~Bulent Sariyildiz, Noe Pion, Philippe Weinzaepfel, and
  Diane Larlus.
\newblock Hard negative mixing for contrastive learning.
\newblock In {\em Advances in Neural Information Processing Systems (NeurIPS)},
  volume~33, pages 21798--21809. Curran Associates, Inc., 2020.

\bibitem{Kingma2015AdamSolver}
Diederik~P. Kingma and Jimmy Ba.
\newblock {Adam: A Method for Stochastic Optimization}.
\newblock In {\em International Conference on Learning Representations (ICLR)},
  2015.

\bibitem{krizhevsky2012Imagenet}
Alex Krizhevsky, Ilya Sutskever, and Geoffrey~E Hinton.
\newblock Image{N}et classification with deep convolutional neural networks.
\newblock In {\em Advances in Neural Information Processing Systems (NIPS)},
  volume~25, pages 1097--1105. Curran Associates, Inc., 2012.

\bibitem{Labbe2019RTAB-Map}
Mathieu Labb{\'e} and Fran{\c c}ois Michaud.
\newblock {RTAB-Map} as an open-source lidar and visual simultaneous
  localization and mapping library for large-scale and long-term online
  operation.
\newblock {\em Journal of Field Robotics}, 36(2):416--446, 2019.

\bibitem{Laskar2020DensePixelMatching}
Zakaria Laskar, Iaroslav Melekhov, Hamed~Rezazadegan Tavakoli, Juha Ylioinas,
  and Juho Kannala.
\newblock Geometric image correspondence verification by dense pixel matching.
\newblock In {\em Proceedings of the IEEE Winter Conference on Applications of
  Computer Vision (WACV)}, pages 2510--2519, 2020.

\bibitem{Lenc2019Detector}
Karel Lenc and Andrea Vedaldi.
\newblock Learning covariant feature detectors.
\newblock In {\em Proceedings of the European Conference on Computer Vision
  (ECCV)}, pages 100--117. Springer International Publishing, 2016.

\bibitem{Li2018Stylization}
Yijun Li, Ming-Yu Liu, Xueting Li, Ming-Hsuan Yang, and Jan Kautz.
\newblock A closed-form solution to photorealistic image stylization.
\newblock In {\em Proceedings of the European Conference on Computer Vision
  (ECCV)}, pages 468--483. Springer International Publishing, 2018.

\bibitem{Li2018Megadepth}
Zhengqi Li and Noah Snavely.
\newblock {MegaDepth}: Learning single-view depth prediction from internet
  photos.
\newblock In {\em Proceedings of the IEEE/CVF Conference on Computer Vision and
  Pattern Recognition (CVPR)}, pages 2041--2050, 2018.

\bibitem{Lowe2004SIFT}
David~G. Lowe.
\newblock Distinctive image features from scale-invariant keypoints.
\newblock {\em International Journal on Computer Vision (IJCV)}, 60(2):91--110,
  2004.

\bibitem{Luo2019Contextdesc}
Zixin Luo, Tianwei Shen, Lei Zhou, Jiahui Zhang, Yao Yao, Shiwei Li, Tian Fang,
  and Long Quan.
\newblock {ContextDesc}: Local descriptor augmentation with cross-modality
  context.
\newblock In {\em Proceedings of the IEEE/CVF Conference on Computer Vision and
  Pattern Recognition (CVPR)}, pages 2522--2531, 2019.

\bibitem{Luo2020ASLFeat}
Zixin Luo, Lei Zhou, Xuyang Bai, Hongkai Chen, Jiahui Zhang, Yao Yao, Shiwei
  Li, Tian Fang, and Long Quan.
\newblock {ASLFeat}: Learning local features of accurate shape and
  localization.
\newblock In {\em Proceedings of the IEEE/CVF Conference on Computer Vision and
  Pattern Recognition (CVPR)}, pages 6588--6597, 2020.

\bibitem{Melekhov2020Stylization}
Iaroslav Melekhov, Gabriel~J. Brostow, Juho Kannala, and Daniyar
  Turmukhambetov.
\newblock Image stylization for robust features.
\newblock {\em ArXiv preprint arXiv:2008.06959}, 2020.

\bibitem{dgcnet}
Iaroslav Melekhov, Aleksei Tiulpin, Torsten Sattler, Marc Pollefeys, Esa Rahtu,
  and Juho Kannala.
\newblock {DGC-Net}: Dense geometric correspondence network.
\newblock In {\em Proceedings of the IEEE Winter Conference on Applications of
  Computer Vision (WACV)}, pages 1034--1042, 2019.

\bibitem{Mishchuk2017HardNet}
Anastasiia Mishchuk, Dmytro Mishkin, Filip Radenovic, and Jiri Matas.
\newblock Working hard to know your neighbor\textquotesingle s margins: Local
  descriptor learning loss.
\newblock In {\em Advances in Neural Information Processing Systems (NIPS)},
  volume~30, pages 4826--4837. Curran Associates, Inc., 2017.

\bibitem{Mishkin2018AffNet}
Dmytro Mishkin, Filip Radenovi{\'c}, and Jiri Matas.
\newblock Repeatability is not enough: Learning affine regions via
  discriminability.
\newblock In {\em Proceedings of the European Conference on Computer Vision
  (ECCV)}, pages 287--304. Springer International Publishing, 2018.

\bibitem{mohan2020efficientps}
Rohit Mohan and Abhinav Valada.
\newblock Efficient{PS}: Efficient panoptic segmentation.
\newblock {\em International Journal on Computer Vision (IJCV)},
  129(5):1551--1579, 2020.

\bibitem{Mueller2019Image2Image}
Markus~S. Mueller, Thorsten Sattler, Marc Pollefeys, and Boris Jutzi.
\newblock Image-to-image translation for enhanced feature matching, image
  retrieval and visual localization.
\newblock {\em ISPRS annals}, IV-2/W7:111--119, 2019.

\bibitem{Mur2017ORB2}
Ra\'ul Mur-Artal and Juan~D. Tard\'os.
\newblock {ORB-SLAM2}: an open-source {SLAM} system for monocular, stereo and
  {RGB-D} cameras.
\newblock {\em IEEE Transactions on Robotics}, 33(5):1255--1262, 2017.

\bibitem{noroozi_2016}
Mehdi Noroozi and Paolo Favaro.
\newblock Unsupervised learning of visual representations by solving jigsaw
  puzzles.
\newblock In {\em Proceedings of the European Conference on Computer Vision
  (ECCV)}, pages 69--84. Springer International Publishing, 2016.

\bibitem{Ono2018LFNet}
Yuki Ono, Eduard Trulls, Pascal Fua, and Kwang~Moo Yi.
\newblock {LF-Net}: Learning local features from images.
\newblock In {\em Advances in Neural Information Processing Systems (NeurIPS)},
  volume~31, pages 6234--6244. Curran Associates, Inc., 2018.

\bibitem{Pautrat2020LISRD}
R{\'{e}}mi Pautrat, Viktor Larsson, Martin~R. Oswald, and Marc Pollefeys.
\newblock Online invariance selection for local feature descriptors.
\newblock In {\em Proceedings of the European Conference on Computer Vision
  (ECCV)}, pages 707--724. Springer International Publishing, 2020.

\bibitem{Porav2018Gans}
Horia Porav, Maddern Will, and Paul Newman.
\newblock Adversarial training for adverse conditions: Robust metric
  localisation using appearance transfer.
\newblock In {\em International Conference on Robotics and Automation (ICRA)},
  pages 1011--1018, 2018.

\bibitem{Pultar2019Amos}
Milan Pultar, Dmytro Mishkin, and Jiri Matas.
\newblock {Leveraging outdoor webcams for local descriptor learning}.
\newblock {\em ArXiv preprint arXiv:1901.09780}, 2020.

\bibitem{Radenovic2018Revisited}
Filip Radenovi{\'c}, Ahmet Iscen, Giorgos Tolias, Yannis Avrithis, and Ond{\v
  r}ei Chum.
\newblock Revisiting oxford and paris: Large-scale image retrieval
  benchmarking.
\newblock In {\em Proceedings of the IEEE/CVF Conference on Computer Vision and
  Pattern Recognition (CVPR)}, pages 5706--5715, 2018.

\bibitem{Radenovic2016GEM}
Filip Radenovi{\'c}, Giorgos Tolias, and Ond{\v r}ei Chum.
\newblock {CNN} image retrieval learns from {BoW}: Unsupervised fine-tuning
  with hard examples.
\newblock In {\em Proceedings of the European Conference on Computer Vision
  (ECCV)}, pages 3--20. Springer International Publishing, 2016.

\bibitem{Radenovic2018TPAMI}
Filip Radenovi{\'c}, Giorgos Tolias, and Ond{\v{r}}ej Chum.
\newblock Fine-tuning {CNN} image retrieval with no human annotation.
\newblock {\em IEEE Transactions on Pattern Analysis and Machine Intelligence},
  41(7):1655--1668, 2019.

\bibitem{Revaud2019R2D2}
Jerome Revaud, Cesar De~Souza, Martin Humenberger, and Philippe Weinzaepfel.
\newblock {R2D2}: Reliable and repeatable detector and descriptor.
\newblock In {\em Advances in Neural Information Processing Systems (NeurIPS)},
  volume~32, pages 12405--12415. Curran Associates, Inc., 2019.

\bibitem{robinson_2021}
Joshua~David Robinson, Ching-Yao Chuang, Suvrit Sra, and Stefanie Jegelka.
\newblock Contrastive learning with hard negative samples.
\newblock In {\em International Conference on Learning Representations (ICLR)},
  2021.

\bibitem{rocco_17}
Ignacio Rocco, Relja Arandjelovi\'c, and Josef Sivic.
\newblock Convolutional neural network architecture for geometric matching.
\newblock In {\em Proceedings of the IEEE/CVF Conference on Computer Vision and
  Pattern Recognition (CVPR)}, pages 2553--2567, 2017.

\bibitem{Rublee2011ORB}
Ethan Rublee, Vincent Rabaud, Kurt Konolige, and Gary Bradski.
\newblock {ORB}: An efficient alternative to {SIFT} or {SURF}.
\newblock In {\em Proceedings of the IEEE/CVF International Conference on
  Computer Vision (ICCV)}, pages 2564--2571, 2011.

\bibitem{sarlin2019coarse}
Paul-Edouard Sarlin, Cesar Cadena, Roland Siegwart, and Marcin Dymczyk.
\newblock From coarse to fine: Robust hierarchical localization at large scale.
\newblock In {\em Proceedings of the IEEE/CVF Conference on Computer Vision and
  Pattern Recognition (CVPR)}, pages 12708--12717, 2019.

\bibitem{sarlin2020superglue}
Paul-Edouard Sarlin, Daniel DeTone, Tomasz Malisiewicz, and Andrew Rabinovich.
\newblock {SuperGlue}: Learning feature matching with graph neural networks.
\newblock In {\em Proceedings of the IEEE/CVF Conference on Computer Vision and
  Pattern Recognition (CVPR)}, pages 4937--4946, 2020.

\bibitem{Savinov2017detector}
Nikolay Savinov, Akihito Seki, Lubor Ladicky, Torsten Sattler, and Marc
  Pollefeys.
\newblock Quad-networks: unsupervised learning to rank for interest point
  detection.
\newblock In {\em Proceedings of the IEEE/CVF Conference on Computer Vision and
  Pattern Recognition (CVPR)}, pages 3929--3937, 2017.

\bibitem{schoenberger2016sfm}
Johannes~Lutz Sch\"{o}nberger and Jan-Michael Frahm.
\newblock Structure-from-motion revisited.
\newblock In {\em Proceedings of the IEEE/CVF Conference on Computer Vision and
  Pattern Recognition (CVPR)}, pages 4104--4113, 2016.

\bibitem{schoenberger2016mvs}
Johannes~Lutz Sch\"{o}nberger, Enliang Zheng, Marc Pollefeys, and Jan-Michael
  Frahm.
\newblock Pixelwise view selection for unstructured multi-view stereo.
\newblock In {\em Proceedings of the European Conference on Computer Vision
  (ECCV)}, pages 501--518. Springer International Publishing, 2016.

\bibitem{Sun2021LoFTR}
Jiaming Sun, Zehong Shen, Yuang Wang, Hujun Bao, and Zhou Xiaowei.
\newblock {LoFTR}: Detector-free local feature matching with transformers.
\newblock In {\em Proceedings of the IEEE/CVF Conference on Computer Vision and
  Pattern Recognition (CVPR)}, pages 8922--8931, 2021.

\bibitem{Taira2018InLoc}
Hajime Taira, Masatoshi Okutomi, Torsten Sattler, Mircea Cimpoi, Marc
  Pollefeys, Josef Sivic, Tomas Pajdla, and Akihiko Torii.
\newblock {InLoc}: Indoor visual localization with dense matching and view
  synthesis.
\newblock In {\em Proceedings of the IEEE/CVF Conference on Computer Vision and
  Pattern Recognition (CVPR)}, pages 1293--1307, 2018.

\bibitem{tang2020NeuralRejection}
Jiexiong Tang, Rares Ambrus, Vitor Guizilini, and Hanme Kim.
\newblock Neural outlier rejection for self-supervised keypoint learning.
\newblock In {\em Proceedings of the International Conference on Machine
  Learning (ICML)}, 2020.

\bibitem{Tian2007L2Net}
Yurun Tian, Bin Fan, and Fuchao Wu.
\newblock {L2-Net}: Deep learning of discriminative patch descriptor in
  euclidean space.
\newblock In {\em Proceedings of the IEEE/CVF Conference on Computer Vision and
  Pattern Recognition (CVPR)}, pages 6128--6136, 2017.

\bibitem{Tian2019SOSNET}
Yurun Tian, Xin Yu, Bin Fan, Fuchao Wu, Huub Heijnen, and Vassileios Balntas.
\newblock {SOSNet}: Second order similarity regularization for local descriptor
  learning.
\newblock In {\em Proceedings of the IEEE/CVF Conference on Computer Vision and
  Pattern Recognition (CVPR)}, pages 11008--11017, 2019.

\bibitem{Torii2015Tokyo247}
Akihiko Torii, Relja Arandjelovi{\'c}, Josef Sivic, Masatoshi Okutomi, and
  Tomas Pajdla.
\newblock 24/7 place recognition by view synthesis.
\newblock In {\em Proceedings of the IEEE/CVF Conference on Computer Vision and
  Pattern Recognition (CVPR)}, pages 257--271, 2015.

\bibitem{Truong2019GLAMpoints}
Prune Truong, Stefanos Apostolopoulos, Agata Mosinska, Samuel Stucky, Carlos
  Ciller, and Sandro~De Zanet.
\newblock {GLAMpoints}: Greedily learned accurate match points.
\newblock In {\em Proceedings of the IEEE/CVF International Conference on
  Computer Vision (ICCV)}, pages 10731--10740, 2019.

\bibitem{Tyszkiewicz2020DISK}
Micha{\l} Tyszkiewicz, Pascal Fua, and Eduard Trulls.
\newblock {DISK}: Learning local features with policy gradient.
\newblock In {\em Advances in Neural Information Processing Systems},
  volume~33, pages 14254--14265. Curran Associates, Inc., 2020.

\bibitem{Verdie2015TILDE}
Yannick Verdie, Kwang~Moo Yi, Vincent Lepetit, and Pascal Fua.
\newblock {TILDE}: A temporally invariant learned detector.
\newblock In {\em Proceedings of the IEEE/CVF Conference on Computer Vision and
  Pattern Recognition (CVPR)}, pages 5279--5288, 2015.

\bibitem{wang2021p2net}
Bing Wang, Changhao Chen, Zhaopeng Cui, Jie Qin, Chris~Xiaoxuan Lu, Zhengdi Yu,
  Peijun Zhao, Zhen Dong, Fan Zhu, Niki Trigoni, and Andrew Markham.
\newblock P2-net: Joint description and detection of local features for pixel
  and point matching.
\newblock In {\em Proceedings of the IEEE/CVF International Conference on
  Computer Vision (ICCV)}, pages 16004--16013, 2021.

\bibitem{Wang2020CAPS}
Qianqian Wang, Xiaowei Zhou, Bharath Hariharan, and Noah Snavely.
\newblock Learning feature descriptors using camera pose supervision.
\newblock In {\em Proceedings of the European Conference on Computer Vision
  (ECCV)}, pages 757--774. Springer International Publishing, 2020.

\bibitem{wu2019detectron2}
Yuxin Wu, Alexander Kirillov, Francisco Massa, Wan-Yen Lo, and Ross Girshick.
\newblock Detectron2.
\newblock \url{https://github.com/facebookresearch/detectron2}, 2019.

\bibitem{Yang2019EPIPOLAR}
Guandao Yang, Tomasz Malisiewicz, and Serge Belongie.
\newblock Learning data-adaptive interest points through epipolar adaptation.
\newblock In {\em Proceedings of the IEEE/CVF Conference on Computer Vision and
  Pattern Recognition (CVPR) Workshops}, pages 1--7, 2019.

\bibitem{Heng2021GeomPerc}
Heng Yang, Wei Dong, Luca Carlone, and Vladlen Koltun.
\newblock Self-supervised geometric perception.
\newblock In {\em Proceedings of the IEEE/CVF Conference on Computer Vision and
  Pattern Recognition (CVPR)}, pages 14350--14361, 2021.

\bibitem{Yang2020UR2KID}
Tsun-Yi Yang, Duy-Kien Nguyen, Huub Heijnen, and Vassileios Balntas.
\newblock {UR2KiD}: Unifying retrieval, keypoint detection, and keypoint
  description without local correspondence supervision.
\newblock {\em ArXiv preprint arXiv:2001.07252}, 2020.

\bibitem{Heng2019MONET}
Yuan Yao, Yasamin Jafarian, and Hyun~Soo Park.
\newblock {MONET}: Multiview semi-supervised keypoint detection via epipolar
  divergence.
\newblock In {\em Proceedings of the IEEE/CVF International Conference on
  Computer Vision (ICCV)}, pages 753--762, 2019.

\bibitem{Yi2016LIFT}
Kwang~Moo Yi, Eduard Trulls, Vincent Lepetit, and Pascal Fua.
\newblock {LIFT}: Learned invariant feature transform.
\newblock In {\em Proceedings of the European Conference on Computer Vision
  (ECCV)}, pages 467--483. Springer International Publishing, 2016.

\bibitem{zhang_2016}
Richard Zhang, Phillip Isola, and Alexei~A. Efros.
\newblock Colorful image colorization.
\newblock In {\em Proceedings of the European Conference on Computer Vision
  (ECCV)}, pages 649--666. Springer International Publishing, 2016.

\bibitem{zhang2017detector}
Xu Zhang, Felix~X. Yu, Svebor Karaman, and Shih-Fu Chang.
\newblock Learning discriminative and transformation covariant local feature
  detectors.
\newblock In {\em Proceedings of the IEEE/CVF Conference on Computer Vision and
  Pattern Recognition (CVPR)}, pages 4923--4931, 2017.

\bibitem{Zhang2020AachenV11}
Zichao Zhang, Torsten Sattler, and Davide Scaramuzza.
\newblock Reference pose generation for long-term visual localization via
  learned features and view synthesis.
\newblock {\em International Journal on Computer Vision (IJCV)},
  129(4):821--844, 2021.

\bibitem{Zhou2021Patch2Pix}
Qunjie Zhou, Torsten Sattler, and Laura Leal-Taix{\'e}.
\newblock {Patch2Pix}: Epipolar-guided pixel-level correspondences.
\newblock In {\em Proceedings of the IEEE/CVF Conference on Computer Vision and
  Pattern Recognition (CVPR)}, pages 4669--4678, 2021.

\bibitem{Zhu2017CycleGAN}
Jun-Yan Zhu, Taesung Park, Phillip Isola, and Alexei Efros.
\newblock Unpaired image-to-image translation using cycle-consistent
  adversarial networks.
\newblock In {\em Proceedings of the IEEE/CVF International Conference on
  Computer Vision (ICCV)}, pages 2242--2251, 2017.

\end{thebibliography}
